\PassOptionsToPackage{numbers, compress}{natbib}
\documentclass[]{antgroup}
\usepackage{antgroup}
\newif\ifshowrevisions
\showrevisionsfalse

\usepackage{pdfrender}

\definecolor{DuoBlue}{HTML}{2D69A3}
\definecolor{DuoMid}{HTML}{328199}
\definecolor{DuoTeal}{HTML}{37988F}

\DeclareRobustCommand{\DuoTitle}{%
\textpdfrender{TextRenderingMode=FillStroke,LineWidth=1.5pt}{%
    \textcolor{DuoBlue}{D}%
    \textcolor{DuoMid}{u}%
    \textcolor{DuoTeal}{o}%
  }%
}

\newcommand{\methodname}{\texttt{\textbf{Duo}} \texttt{\textbf{v1}}\xspace}

\newcommand{\papertitle}{\DuoTitle: A Reciprocal World Model for Full-Duplex Interaction}
\newcommand{\wrm}{\texttt{Reciprocal} \texttt{World} \texttt{Model}\xspace}
\newcommand{\rwmshort}{\texttt{RWM}\xspace}

\newcommand{\tocite}[1]{{\color{red} [TO CITE]}}

\definecolor{CQColor}{rgb}{0.0,0.0,1.0} 

\definecolor{TSColor}{rgb}{0.5,0.0,0.8} 

\definecolor{CQRColor}{rgb}{1.0,0.0,1.0} 

\usepackage{graphicx}
\usepackage{xspace}
\usepackage{tikz}
\usepackage{todonotes}
\usepackage{multirow}
\usepackage{amsmath}
\usepackage{cleveref}
\usepackage{subcaption}
\usepackage{multicol}
\usepackage{amssymb}
\usepackage{booktabs}
\usepackage{array}
\usepackage{bm}
\usepackage{enumitem}
\usepackage{CJKutf8}
\usepackage{algorithm}
\usepackage{algpseudocode}
\usepackage{tabularx}
\usepackage{bbm}
\usepackage{makecell}
\usepackage{threeparttable}
\usepackage{float}
\usepackage{wrapfig}
\usepackage[normalem]{ulem}
\useunder{\uline}{\ul}{}

\usepackage{amsmath,amsfonts,bm}

\def\figref#1{figure~\ref{#1}}

\def\secref#1{section~\ref{#1}}

\def\eqref#1{equation~\ref{#1}}

\def\algref#1{algorithm~\ref{#1}}

\def\1{\bm{1}}

\DeclareMathAlphabet{\mathsfit}{\encodingdefault}{\sfdefault}{m}{sl}
\SetMathAlphabet{\mathsfit}{bold}{\encodingdefault}{\sfdefault}{bx}{n}

\newcommand*\justify{%
  \fontdimen2\font=0.4em%
  \fontdimen3\font=0.2em%
  \fontdimen4\font=0.1em%
  \fontdimen7\font=0.1em%
  \hyphenchar\font=`\-%
}

\renewcommand{\texttt}[1]{%
  \begingroup
  \ttfamily
  \begingroup\lccode`~=`/\lowercase{\endgroup\def~}{/\discretionary{}{}{}}%
  \begingroup\lccode`~=`[\lowercase{\endgroup\def~}{[\discretionary{}{}{}}%
  \begingroup\lccode`~=`.\lowercase{\endgroup\def~}{.\discretionary{}{}{}}%
  \catcode`/=\active\catcode`[=\active\catcode`.=\active
  \justify\scantokens{#1\noexpand}%
  \endgroup
}

\usepackage{amsmath}
\usepackage{amssymb}
\usepackage{amsfonts}                               %
\usepackage{amsthm}
\usepackage[mathcal]{eucal}
\usepackage{mathrsfs}
\usepackage{bm}                                     %
\usepackage{blkarray}                               %
\usepackage{nicefrac}                               %

\usepackage{wrapfig}
\usepackage{graphicx}                               %
\usepackage{caption}
\usepackage{cleveref}

\usepackage{tikz}                                          
\usepackage{circuitikz}
\usetikzlibrary{patterns}
\usetikzlibrary{positioning,calc,fit,decorations.pathmorphing,shapes.geometric, shapes.gates.logic.US, calc}
\usetikzlibrary{arrows,arrows.meta,decorations.markings,shapes,shapes.arrows}
\usetikzlibrary{decorations,decorations.pathreplacing}
\usetikzlibrary{backgrounds}
\usepackage{pgfplots}
\usepackage{pgfplotstable}
\usepgfplotslibrary{groupplots}
\usepackage{scalefnt}
\pgfplotsset{compat=newest}

\makeatletter
\patchcmd{\@addtotoporbot}
  {\@midlist\@botlist}
  {\@toplist\@midlist\@botlist}
  {}{\PackageError{manuscript}{Cannot enforce one top float per type}{}}
\makeatother

\definecolor{firstcolor}{HTML}{C3423F}
\definecolor{secondcolor}{HTML}{2A4B8C}
\definecolor{myblue}{HTML}{D9E6FF}
\definecolor{statusgreen}{HTML}{26734D}
\definecolor{statusorange}{HTML}{CC7000}

\usepackage{tocloft}
\newenvironment{talign*}
  {\csname align*\endcsname}
  {\endalign}

\definecolor{darkred}{RGB}{150,0,0}
\providecommand{\darkredtext}[1]{\textcolor{darkred}{#1}}

\renewcommand{\eqref}[1]{%
  \hyperref[#1]{\darkredtext{\textup{Eq.~(\ref*{#1})}}}}
\renewcommand{\secref}[1]{%
  \hyperref[#1]{\darkredtext{Sec.~\ref*{#1}}}}
\renewcommand{\figref}[1]{%
  \hyperref[#1]{\darkredtext{Fig.~\ref*{#1}}}}
\renewcommand{\algref}[1]{%
  \hyperref[#1]{\darkredtext{Alg.~\ref*{#1}}}}

\providecommand{\thmref}[1]{\hyperref[#1]{\darkredtext{Thm.~\ref*{#1}}}}
\providecommand{\defref}[1]{\hyperref[#1]{\darkredtext{Def.~\ref*{#1}}}}
\providecommand{\propref}[1]{\hyperref[#1]{\darkredtext{Prop.~\ref*{#1}}}}
\providecommand{\assumpref}[1]{\hyperref[#1]{\darkredtext{Assump.~\ref*{#1}}}}
\providecommand{\remarkref}[1]{\hyperref[#1]{\darkredtext{Rem.~\ref*{#1}}}}
\providecommand{\hypref}[1]{\hyperref[#1]{\darkredtext{Hyp.~\ref*{#1}}}}
\providecommand{\conjref}[1]{\hyperref[#1]{\darkredtext{Conj.~\ref*{#1}}}}
\providecommand{\lemref}[1]{\hyperref[#1]{\darkredtext{Lem.~\ref*{#1}}}}
\providecommand{\corref}[1]{\hyperref[#1]{\darkredtext{Cor.~\ref*{#1}}}}
\providecommand{\noteref}[1]{\hyperref[#1]{\darkredtext{Nota.~\ref*{#1}}}}
\providecommand{\claimref}[1]{\hyperref[#1]{\darkredtext{Clm.~\ref*{#1}}}}
\providecommand{\obsref}[1]{\hyperref[#1]{\darkredtext{Obs.~\ref*{#1}}}}
\providecommand{\tabref}[1]{%
  \hyperref[#1]{\darkredtext{Tab.~\ref*{#1}}}}
\providecommand{\appref}[1]{%
  \hyperref[#1]{\darkredtext{App.~\ref*{#1}}}}

\theoremstyle{plain}

\theoremstyle{definition}

\theoremstyle{remark}

\tcbset{
  llada theorem box/.style={
    enhanced,
    breakable,
    sharp corners,
    boxrule=0pt,
    leftrule=2.5pt,
    colback=title_blue!4!white,
    colframe=title_blue,
    left=6pt,
    right=6pt,
    top=3pt,
    bottom=3pt,
    before skip=7pt,
    after skip=7pt
  },
  llada remark box/.style={
    llada theorem box,
    colback=statusorange!5!white,
    colframe=statusorange
  }
}

\tcolorboxenvironment{theorem}{llada theorem box}
\tcolorboxenvironment{proposition}{llada theorem box}
\tcolorboxenvironment{lemma}{llada theorem box}
\tcolorboxenvironment{corollary}{llada theorem box}
\tcolorboxenvironment{conjecture}{llada theorem box}
\tcolorboxenvironment{definition}{llada theorem box}
\tcolorboxenvironment{assumption}{llada theorem box}
\tcolorboxenvironment{hypothesis}{llada theorem box}
\tcolorboxenvironment{notation}{llada theorem box}
\tcolorboxenvironment{claim}{llada theorem box}
\tcolorboxenvironment{problem}{llada theorem box}
\tcolorboxenvironment{observation}{llada theorem box}
\tcolorboxenvironment{remark}{llada remark box}

\usepackage{bookmark}
\usepackage{needspace}
\usepackage{fontawesome5}
\hypersetup{pdftitle={Towards World Response Models for Real-Time Co-Presence},pdfauthor={Author 1, Author 2, Author 3, Author 4, Author 5, Author 6, Author 7, Author 8, Author 9, Author 10, Author 11, Author 12, Author 13, Author 14, Author 15},pdfsubject={Streaming audio-visual generation and human-character interaction}}
\newcolumntype{Y}{>{\raggedright\arraybackslash}X}
\providecommand{\code}[1]{{\ttfamily\detokenize{#1}}}
\patchcmd{\mymaketitle}{\vskip 3mm}{\vskip 1mm}{}%
  {\PackageError{duo}{Cannot adjust author contribution spacing}{}}
\patchcmd{\mymaketitle}{\vskip 6mm}{\vskip 3mm}{}%
  {\PackageError{duo}{Cannot adjust author top spacing}{}}
\patchcmd{\maketitle}{\vskip 4mm}{\vskip 2mm}{}%
  {\PackageError{duo}{Cannot adjust title bottom spacing}{}}

\newsavebox{\duoheaderant}
\newcommand{\duoheaderbrands}{%
  \sbox{\duoheaderant}{%
    \includegraphics[width=0.1\linewidth]{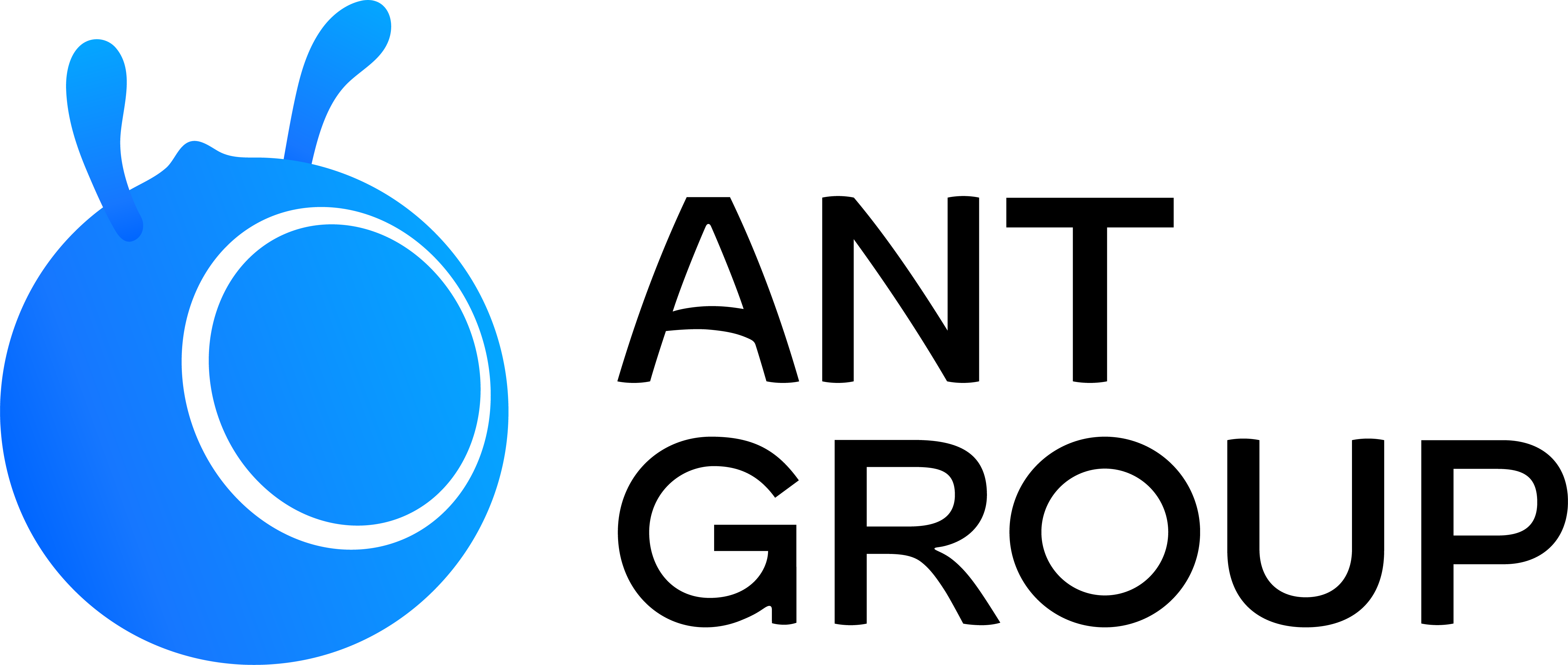}}%
  \usebox{\duoheaderant}\hspace{4mm}%
  \raisebox{0.005\ht\duoheaderant}[0pt][0pt]{%
    \includegraphics[height=0.99\ht\duoheaderant]{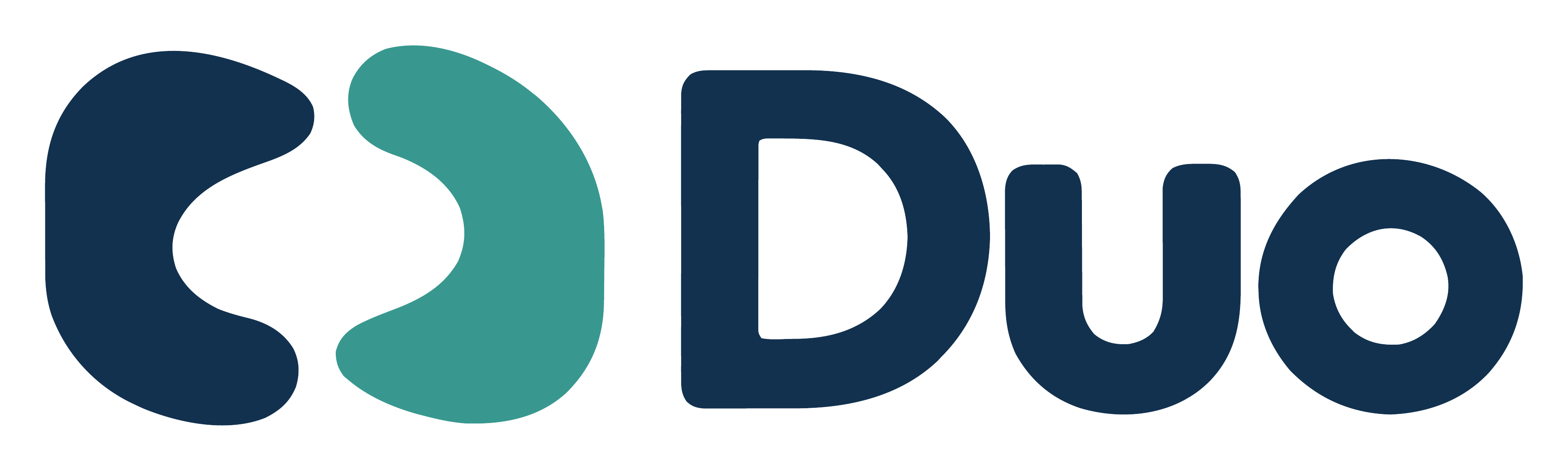}}%
}
\fancypagestyle{firststyle}{%
  \fancyhead[R]{\sffamily\fontsize{9}{11}\selectfont\color{black}Technical Report}%
  \fancyhead[L]{\vskip 4.5mm\duoheaderbrands}%
}

\title{\papertitle}
\renewcommand{\authorlist}{%
  \authorformat{Hengyuan Cao}, 
  \authorformat[\dagger]{Biao Gong},
  \authorformat{Guoshan Lu}, 
  \authorformat{Qiyuan Zhang},
  \authorformat{Lin Liu}, \authorformat{Yiding Tian},
  \authorformat{Yilei Jiang},\\ \authorformat{Yuxin Ma},
  \authorformat{Chengzhang Wu}, \authorformat{Kun Chen},
  \authorformat{Yiqian Xie}, 
  \authorformat{Ting Liu}, 
  \authorformat{Xin Zheng},
  \authorformat{Xijun Gu},
  \authorformat{Tianren Ma},
  \authorformat{Jiancheng Pan},
  \authorformat{Chi Zhang},
  \authorformat{Xuzhe Zheng},
  \authorformat{Siyi Li},
  \authorformat[\dagger]{Zhenzhong Lan}%
}
\contribution[\dagger]{Project Lead}
\begin{document}
\maketitle
\vspace{3mm}
\begin{abstract}
In this report, we introduce the \wrm (\rwmshort) and present \methodname,
a full-duplex world model that continuously perceives audiovisual inputs while responding to users in real time.
The \rwmshort treats intelligence and physical reality as equally fundamental to what constitutes a world. Existing world models largely focus on simulating the latter, leaving reasoning and decision-making as external controls rather than intrinsic parts of the world itself.
In \rwmshort, we identify two equally fundamental capabilities: 
\emph{Intent}, which provides the cognition underlying world interactions, and \emph{Realization}, which generates coherent audiovisual behavior.
To realize this formulation, \methodname introduces three key technical designs: (1) Leveraging autoregressive modeling for continuous interaction reasoning and participation decisions, rather than restricting it to instruction understanding for diffusion models. 
(2) Constructing chunk-specific conditions and synchronize tokens across Intent and Realization on a unified timeline. Module-specific attention masks enforce causal dependencies both within and across chunks, preventing access to future inputs.
(3) Mitigating distribution shift in both modules through dual rollout distillation, training on autoregressively generated Intent tokens and synthesized audiovisual histories from Realization.
Both modules continuously perceive real-time audio, video, and text, enabling full-duplex interaction with spontaneous conversations and interruptions.
Beyond conversation, users can share real objects, reshape virtual scenes, guide assistants, or play audiovisual games without mouse or keyboard controls.
The release of \methodname makes our first step, with the next version coming soon. We look forward to broader exploration of this emerging field.
\end{abstract}

\newcommand{\duohomeurl}{https://duo-v1.github.io}
\newcommand{\duogithuburl}{https://github.com/inclusionAI/Duo}
\newcommand{\duohfurl}{https://huggingface.co/collections/inclusionAI/duo-v1}

\vspace{-3mm}
\begingroup
\setlength{\tabcolsep}{0pt}
\renewcommand{\arraystretch}{1.05}
\noindent\makebox[\linewidth][c]{%
\begin{tabular}{@{}p{13pt}@{\hspace{6pt}}p{110pt}@{\hspace{10pt}}l@{}}
  \href{\duohomeurl}{\raisebox{-1.7pt}{\includegraphics[width=13pt]{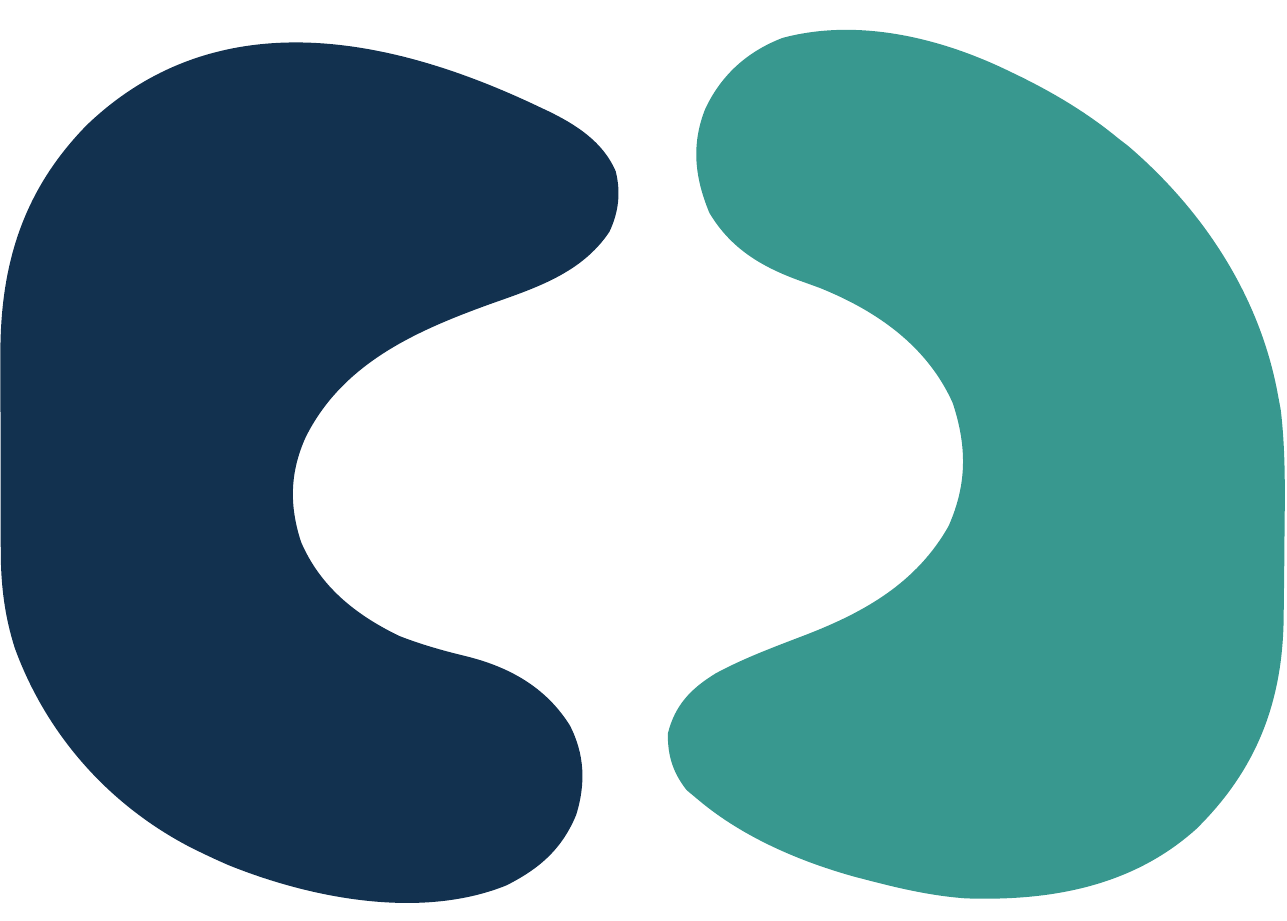}}}
  & \textbf{Homepage}
  & {\normalsize\expandafter\url\expandafter{\duohomeurl}} \\[2pt]
  \raisebox{-0.7pt}{\large\faGithub}
  & \textbf{GitHub}
  & {\normalsize\expandafter\url\expandafter{\duogithuburl}} \\[2pt]
  \raisebox{-2.1pt}{\includegraphics[height=11pt]{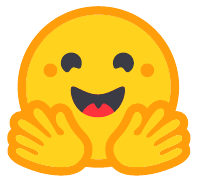}}
  & \textbf{Hugging Face Model}
  & {\normalsize\expandafter\url\expandafter{\duohfurl}}
\end{tabular}
}
\par
\endgroup
\vspace{2pt}

\begingroup
\setlength{\intextsep}{1pt}
\begin{figure}[H]
  \centering
  \scalebox{0.98}{
  \includegraphics[width=\linewidth]{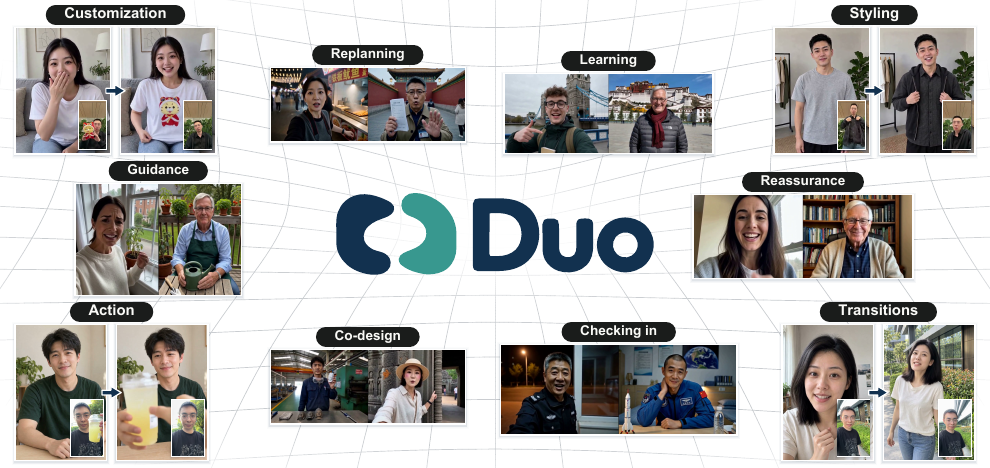}
  }
  \vspace{-2mm}
  \caption{Illustrative scenarios for full-duplex interaction with \wrm.}
  \vspace{-2mm}
  \label{fig:teaser}
\end{figure}

\endgroup
\clearpage
\setcounter{tocdepth}{3}
\tableofcontents
\clearpage
\section{Introduction}
\label{sec:introduction}

World models have made rapid progress in simulating the physical side of a world~\citep{ha2018worldmodels,hafner2025dreamer,brooks2024sora}: given past observations and control signals, they predict how scenes evolve~\citep{bruce2024genie,valevski2025gamengen,hu2023gaia,agarwal2025cosmos}, with increasingly realistic dynamics, appearance, and long-range consistency~\citep{deepmind2024genie2,deepmind2025genie3,robbyant2026lingbot,lingbot-world-v2,chen2026hy,huang2026solarwm}.
In most of these works, however, intelligence resides outside the world.
A user steers the generated world through inputs with limited degrees of freedom, such as keyboard and mouse commands, gamepad actions, or text prompts, but nothing within the world perceives the full state of the user in real time or decides how to respond.
Such worlds can be accurate, yet they remain lifeless: \textit{there is a world to explore, but no one to talk to.}

We argue that intelligence and physical reality are equally fundamental to a world. Building on this view, we introduce the \wrm{} (\rwmshort{}).
The \rwmshort{} generates a world inhabited by living presences and connects it to the real world through full-duplex, real-time interaction. Each side continuously perceives audio and video from the other and can respond, interrupt, or remain silent while the other is still acting.
This reciprocal interaction requires two complementary capabilities: \emph{Intent}, which provides the cognition underlying world interactions, and \emph{Realization}, which generates coherent audiovisual behavior. These functional roles can be fulfilled by a unified multimodal model or by a modular system combining a multimodal language model with a diffusion model. In this report, we present \methodname{}, a modular instantiation of the \rwmshort{}. Interaction proceeds in chunks of 17 video frames, approximately 0.71\,s at 24\,fps. Three technical designs coordinate Intent and  Realization within a single streaming system.

\paragraph{Autoregressive intent for interaction.} In most audio-visual video generative model~\citep{h3modelcard}, a language model serves as an instruction encoder for the diffusion model. In \methodname{}, the intent module instead reasons over the ongoing interaction autoregressively and makes participation decisions: at every chunk, it emits a \texttt{<speak>} action followed by text or a \texttt{<wait>} action, conditioned on the user input so far and its own earlier language. Its hidden states, rather than its decoded text alone, carry each decision to the realization module (\secref{sec:method}). Action-aware rebalancing prevents the intent module from collapsing into perpetual waiting, since most chunks contain no new assistant speech, and history perturbations teach it to recover from imperfect earlier responses (\secref{sec:training}).

\paragraph{Causal chunk conditions on a unified timeline.} User speech, user video, intent tokens, and assistant media are synchronized on one timeline, and each chunk is given exactly the conditions that are available at its decision boundary. Module-specific attention masks enforce causality within and across chunks. In the intent module, temporal patches are completed inside each chunk, so no future frame leaks into a decision through the visual frontend. In the realization module, noisy media tokens of a chunk can attend to the reference, intent features, and clean history, but not to their own clean targets or to any later chunk (\secref{sec:method}).

\paragraph{Dual rollout distillation.} Before distillation, the realization module is trained on intent features of recorded assistant text and on recorded media history, whereas in a live exchange both are produced by the system itself.
Dual rollout brings inference-time conditioning into distillation: within each training window, the frozen Intent Module autoregressively generates the final assistant turn, and the Realization Module conditions on the resulting intent features and its own generated audiovisual history.
Distribution matching distillation on these rollouts yields a four-step streaming generator trained under conditions that match live interaction.

We develop several data construction pipelines and a streaming runtime to support \methodname{}.
The training data combine scripted synthetic video calls and recorded conversations, with audio, video, and text aligned on a shared timeline for causal interaction training (\secref{sec:data}).
The streaming pipeline coordinates intent inference, audiovisual generation, and incremental decoding while continuing to capture user inputs (\secref{sec:runtime}). 
Speech and vision provide a rich, direct channel between the two worlds, conveying information that keyboard and mouse commands often fail to capture, including tone, timing, expression, and visual details of real objects, actions, and scenes.
This richer interface creates opportunities for a wide range of shared activities shaped jointly by real and generated participants.

\section{Related Work}
\label{sec:related}

\paragraph{World models and persistent interaction.}
World Models and Dreamer learn predictive representations for control~\citep{ha2018worldmodels,hafner2025dreamer}, while Genie learns action-controllable environments from video~\citep{bruce2024genie}. LingBot-World studies a data engine and staged causal adaptation for continuing visual worlds~\citep{robbyant2026lingbot}. These works connect learned dynamics to an agent's actions and subsequent observations.

\paragraph{Full-duplex and multimodal participation.}
Moshi represents concurrent user and assistant speech streams~\citep{moshiteam2024moshi}. Qwen-Omni separates multimodal understanding from streaming speech generation through connected Thinker and Talker components~\citep{xu2025qwen25omni,xu2025qwen3omni}. Wan-Streamer extends the streaming setting to audio--visual foundation models and their execution schedule~\citep{huang2026wanstreamer}. ELLSA jointly models full-duplex speech, gesture, and expression, including non-verbal behavior while listening~\citep{wang2026ellsa2}. These works establish participation timing and visible communicative behavior as closely related modeling problems.

\paragraph{Joint and streaming audiovisual generation.}
Ovi, MOVA, and LTX develop synchronized audio--visual generation through coupled modality streams~\citep{low2025ovi,openmoss2026mova,hacohen2026ltx2}. MiniMax-H3 is a reference-conditioned audio--visual generator whose weights can initialize related models~\citep{h3modelcard}. Audio-driven animation methods such as EMO and OmniHuman realize visible behavior from supplied speech~\citep{tian2024emo,lin2025omnihuman}. LiveTalk combines a distilled video performer with streaming audio from Qwen-Omni~\citep{chern2025livetalk}. These approaches connect speech with visible articulation and motion, with different choices of conditioning and generation order. Streaming joint diffusion further requires audio and video to remain aligned as their generated histories grow. OmniForcing addresses this conversion through asymmetric block-causal alignment, audio-sink stabilization, and joint self-generated rollouts~\citep{su2026omniforcing}. OmniLiveAvatar combines progressive autoregressive distillation, synchronized audio-video memory, and rolling prompt planning~\citep{zhu2026omniliveavatar}. 

\paragraph{Causal adaptation and few-step training.}
Diffusion Forcing assigns different noise levels across a sequence, and CausVid distills a bidirectional generator into a causal few-step student~\citep{chen2024diffusionforcing,causvid2025}. Self Forcing uses self-generated history during training to reduce the context mismatch with autoregressive inference~\citep{yin2025selfforcing}. Work on causal-compatible initialization further aligns teacher and student during the transition to few-step generation~\citep{zhu2026causalforcing,zhao2026causalforcingpp}. Consistency models and distribution matching provide complementary tools for reducing sampling steps~\citep{song2023consistency,yin2024dmd}, with later distribution-matching work incorporating multistep alignment and adversarial training~\citep{yin2024dmd2}.

\section{Architecture}
\label{sec:method}

\begin{figure}[t]
  \centering
  \includegraphics[width=\linewidth]{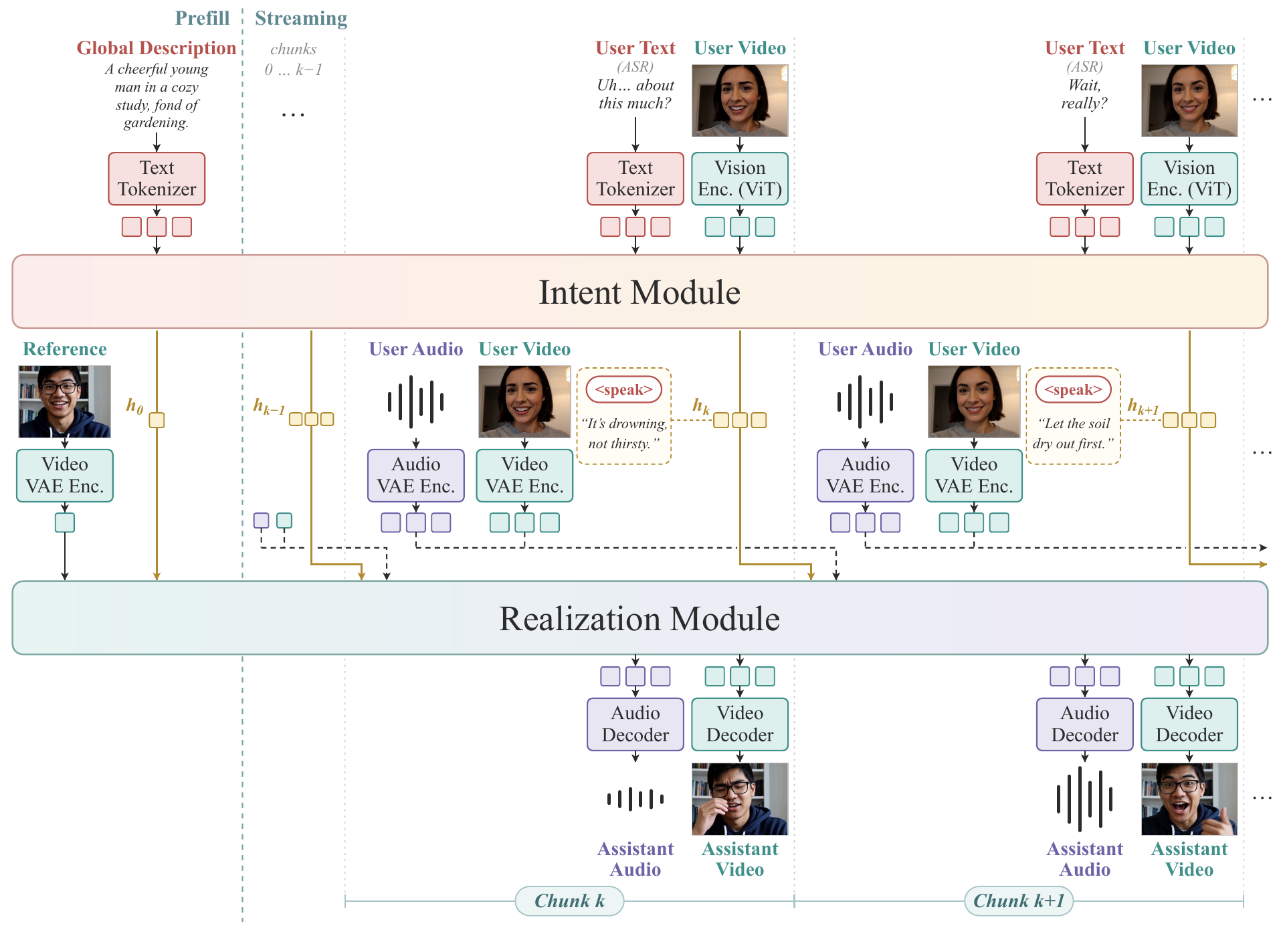}
  \caption{\textbf{The framework of \methodname{}.} A joint causal diffusion transformer is conditioned on hidden states from the Intent Module, directly encoded user video and audio, a visual reference, and previously generated assistant media. The conversation media illustrate an offline, script-conditioned example.}
  \label{fig:architecture}
\end{figure}

\methodname{} couples a language-based \emph{Intent Module} with a causal audiovisual \emph{Realization Module} (\figref{fig:architecture}). Hidden states produced by the Intent Module serve as the interface between the two modules, providing semantic guidance for the joint generation of video and stereo audio for the assistant. The two modules process complementary inputs and maintain separate histories, integrating response planning and audiovisual synthesis within a unified architecture. The following subsections describe this design in detail: \secref{sec:worldresponsemodel} formalizes the interaction between the two modules, \secref{sec:intent} details the Intent Module, \secref{sec:vae} describes the audio-visual latent space, and \secref{sec:joint_dit} presents the causal joint generator used by the Realization Module.

\subsection{The Overview of Reciprocal World Model}
\label{sec:worldresponsemodel}
\label{sec:architecture-overview}
\label{sec:foundations}
\label{sec:reciprocal-formulation}

\rwmshort treats intelligence and physical reality as equally fundamental components of a world. Their interaction is reciprocal: Intent determines how the system responds to the evolving environment, while Realization expresses that intent through audiovisual behavior, which in turn becomes part of the environment for subsequent interaction. We formalize this process as a chunk-wise, full-duplex interaction.

An interaction is divided into consecutive chunks of fixed duration, and $\mathbf{c}_k$ denotes the chunk at time index $k$. A chunk is a time interval and need not align with a word, a sentence, or a conversational turn. For each chunk $\mathbf{c}_k$, let $u_k^v$, $u_k^t$, and $u_k^a$ denote the incoming video, speech transcript, and audio, respectively. Let $y_k$ denote the language output generated by \methodname{}, either \code{<speak>} followed by text or \code{<wait>}. Let $h_k$ denote the hidden states produced by the Intent Module for this output, and $z_k=(z_k^v,z_k^a)$ the latent representations of the generated video and audio. We use the subscript $<k$ to denote the history of a sequence up to, but not including, chunk $\mathbf{c}_k$. $d$ denotes the global instructions, and $I_{\mathrm{ref}}$ denotes the reference image.

At each chunk $k$, \methodname{} generates a response $\mathbf{c}_k$ in two steps, where $<k$ denotes the history before chunk $k$. First, the Intent Module conditions on the global instructions $d$, user video embeddings $x^v_{u,<k}$, user text $u_{t,<k}$, and previous assistant text outputs $y_{<k}$:
\begin{equation}
\begin{aligned}
&C^{\phi}_k = \big(d,\ x^v_{u,<k},\ u_{t,<k},\ y_{<k}\big),\\
&y_k \sim p_\phi\big(\,\cdot \mid C^{\phi}_k\big),
\qquad h_k = f_\phi\big(C^{\phi}_k,\ y_k\big).
\end{aligned}
\label{eq:intent-step}
\end{equation}
The model forward pass produces the text output $y_k$ and hidden states $h_k$, which convey the intent to the Realization Module. The Realization Module jointly generates audio and video latents, conditioned on the reference image $I_{\mathrm{ref}}$, the intent hidden states $h_k$, user video embeddings $x^v_{u,<k}$, user audio $u_{a,<k}$, and previously generated assistant latents $z^a_{<k}$ and $z^v_{<k}$:
\begin{equation}
\begin{aligned}
&C^{\theta}_k = \big(I_{\mathrm{ref}},\ h_k,\ x^v_{u,<k},\ u_{a,<k},\ z^a_{<k},\ z^v_{<k}\big),\\
&\big(z^a_k,\ z^v_k\big) \sim p_\theta\big(\,\cdot \mid C^{\theta}_k\big).
\end{aligned}
\label{eq:realization-step}
\end{equation}
The generated latents $z^a_k$ and $z^v_k$ are decoded into the audio and video of the response chunk $\mathbf{c}_k$.

This formulation has two properties. First, it is causal. Both modules condition on interaction histories preceding chunk $k$, and the Realization Module additionally receives the current intent representation $h_k$. Neither module accesses future inputs. User inputs received during response generation can be incorporated into subsequent chunks, supporting full-duplex interaction. Second, it supports continuity across chunks. The text output $y_k$ and generated latents $z^a_k$ and $z^v_k$ are added to the corresponding histories for subsequent chunks, allowing the system to continue from previous text, audio, and video outputs. \secref{sec:intent} describes how the Intent Module implements $p_\phi$, \secref{sec:vae} describes the audio-visual latent representations $z^a_k$ and $z^v_k$, and \secref{sec:joint_dit} describes how the Realization Module implements $p_\theta$.

\subsection{Multimodal Intent Module}
\label{sec:intent}
\label{sec:observation-adapter}

The Intent Module adapts a pretrained multimodal language backbone~\citep{qwen2025qwen3vl} to causal, chunk-wise interaction. Given the context, it generates the text output $y_k$ and provides hidden states $h_k$.

\paragraph{Causal input assembly.}
Inputs enter $C^\phi_k$ according to when they become available, without requiring all modalities to share a sampling rate. ASR recovers user text $u_t$ from user audio $u_a$, and forced alignment assigns each text token a span on the source audio timeline. A token is admitted only after the end of its aligned span. Video frames and audio samples are mapped to the same half-open chunk intervals using source timestamps.

\begin{figure}[t]
  \centering
  \includegraphics[width=\textwidth]{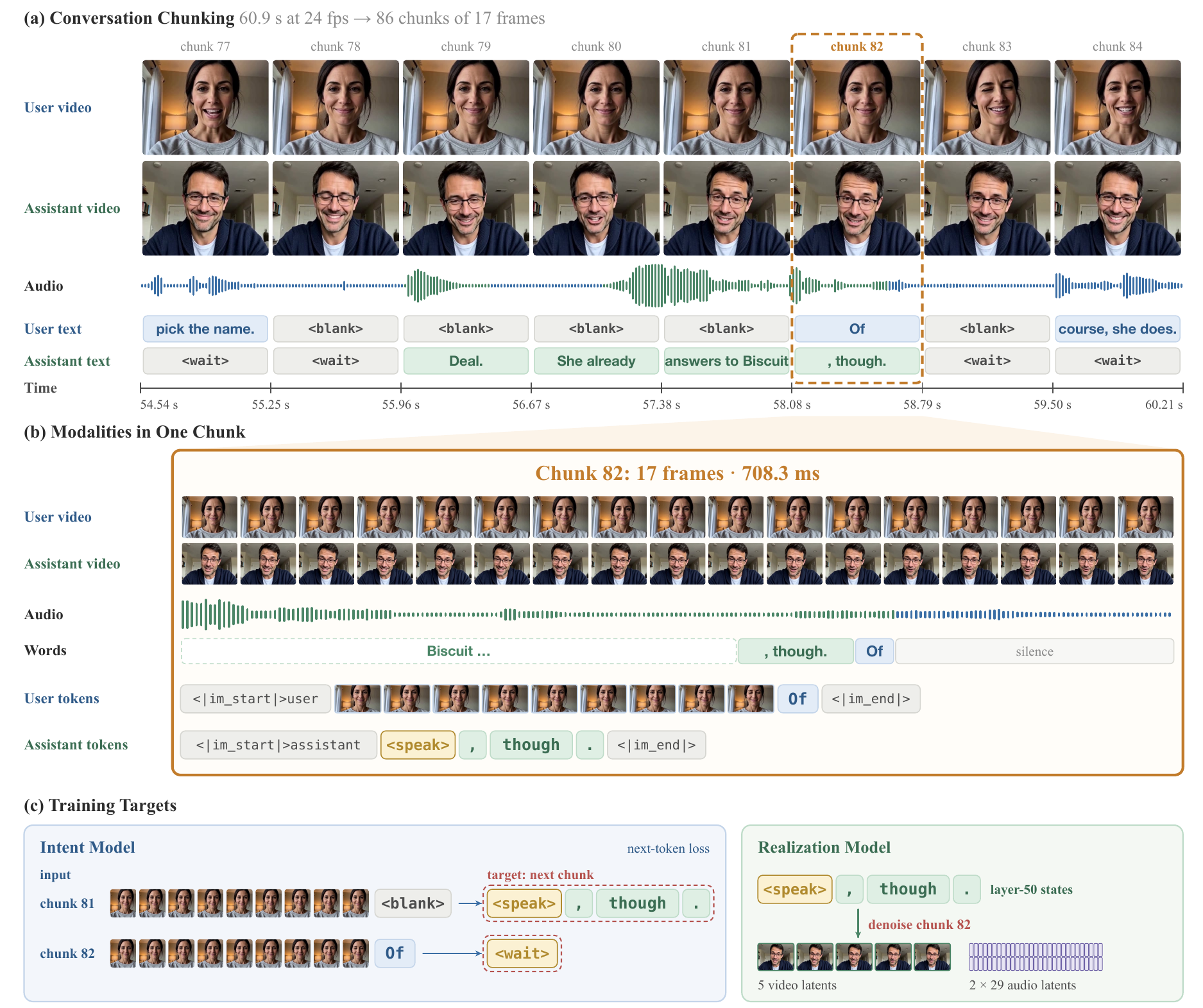}
  \caption{\textbf{ASR-aligned chunk construction and causal supervision.} On a shared audiovisual timeline, user text is assigned to chunks by aligned token end times, while assistant text targets are assigned by speech onset. The token \code{<blank>} marks the absence of newly available user text. An assistant chunk with no new speech onset uses \code{<wait>}, even if speech from the preceding chunk continues. At chunk $k$, the Intent Module predicts $y_k$ from $C^\phi_k$, using user inputs available through chunk $k-1$. The initial output $y_0$ is fixed to \code{<wait>}. Only subsequent assistant targets contribute to the loss.}
  \label{fig:intent-chunk-alignment}
\end{figure}

Assistant text targets are assigned to chunks according to speech onset, while the initial chunk uses a fixed \code{<wait>} action (\figref{fig:intent-chunk-alignment}). We preserve fine-grained alignment before constructing chunk views, avoiding early rounding of speech boundaries. Each input context therefore contains only information available at the corresponding decision boundary, and different chunk configurations can be derived from the same aligned record. The ASR and alignment tools, together with their quality checks, are described in \secref{sec:data-admission}.

\paragraph{Causal visual encoding.}
Causality must also hold before attention is applied. Temporal patch construction can combine multiple source frames, so an attention mask cannot remove future information already mixed into a visual token. We therefore form each temporal group entirely within one input chunk. For a native 17-frame chunk with source frames numbered $0,\ldots,16$, the final pair is $(16,16)$: the last frame is repeated instead of paired with frame $17$ from the next chunk. Each group becomes available at the latest timestamp among its source frames. This preserves the pretrained visual input format while preventing temporal grouping from crossing chunk boundaries.

\paragraph{Participation and language outputs.}
The text output $y_k$ contains either \code{<speak>} followed by a text body or \code{<wait>} without new text. Both actions are generated through the standard token prediction path. The explicit action separates the decision to begin speaking from the words of the response, since appropriate content can still be delivered at the wrong time. A \code{<wait>} output can coexist with ongoing playback and media continuation. Participation supervision, target balancing, and history augmentation are described in \secref{sec:intent-training}.

\subsection{Variational Autoencoder}
\label{sec:vae}

To support real-time response generation, we encode audio and video into compact latent representations $z^a$ and $z^v$. For audio, each stereo channel is compressed into 40 latent vectors per second, with 32 dimensions per vector. For video, the encoder processes groups of 17 frames, with padding applied when fewer than 17 frames are available. Each group consists of one keyframe followed by four packets of four frames, following a $1$-$4$-$4$-$4$-$4$ temporal arrangement. These 17 frames are encoded into 5 latent frames. Each latent frame has 24 dimensions, with spatial resolution reduced by a factor of $16\times16$. For input height $H$ and width $W$, an RGB frame group of shape $(17,H,W,3)$ is therefore encoded into a latent:
\begin{equation}
z_{\mathrm{group}}^v \in \mathbb{R}^{5 \times \frac{H}{16} \times \frac{W}{16} \times 24}.
\end{equation}%
Before entering the diffusion transformer, the video latents undergo spatial merging over $2 \times 2$ patches, producing a video input vector with $2 \times 2 \times 24 = 96$ dimensions. Each audio latent vector has 32 dimensions per channel. The decoding procedure is described in \secref{sec:runtime}.

\subsection{Joint Causal Audiovisual Realization}
\label{sec:joint_dit}
\label{sec:block_causality}

The Realization Module uses a causal audiovisual diffusion transformer initialized with pretrained weights from~\citep{h3modelcard}. Causal adaptation and distillation of the pretrained backbone are described in \secref{sec:mixed_training}. The module jointly generates audio and video in the latent spaces described in \secref{sec:vae}, implementing the conditional generation process in \eqref{eq:realization-step} through flow matching~\citep{lipman2023flowmatching}. Starting from random noise, the model generates audiovisual latents conditioned on the intent representation, reference image, available user inputs, and previously generated assistant media. Access to this history supports continuity across response chunks.

\paragraph{Intent conditioning.}
\label{sec:conditioning-interface}
\label{sec:executed-prefix}
A linear projection maps $h_k$ from the Intent Module to the hidden width of the Realization Module, and a refiner network produces text-conditioning embeddings. An attention mask restricts feature computation to token positions associated with the global instructions $d$ and the current assistant text output $y_k$:
\begin{equation}
x_k^t =
\operatorname{Refine}_\theta\!\left(
\operatorname{Project}_\theta(h_k);
M_k^{\mathrm{feat}}
\right).
\label{eq:conditioning-bridge}
\end{equation}

\paragraph{Media conditioning.}
User audio and video are encoded by the respective audio and video VAE encoders described in \secref{sec:vae}, producing conditioning latents $z^a_{u,<k}$ and $z^v_{u,<k}$. The reference image $I_{\mathrm{ref}}$ is encoded by the video VAE into reference latents $z^r$. These latents are mapped to the corresponding input embeddings $x^a_{u,<k}$, $x^v_{u,<k}$, and $x^r$ for the joint DiT. Text conditioning is provided by $x_k^t$, derived from hidden states associated with the global instructions $d$ and the current assistant output $y_k$, as described above. Previously generated assistant latents $z^a_{<k}$ and $z^v_{<k}$ provide audiovisual history. Together, these conditions supply user observations, reference appearance, semantic guidance, and generated history for the current response chunk. Only user inputs available before chunk $k$ are included.

\paragraph{Positional embedding.}
\label{sec:position_time}
We use 3D-RoPE to represent temporal and spatial relationships among tokens. The text-conditioning embeddings use only the temporal coordinate, with both spatial coordinates set to zero. Audio tokens use the two extreme spatial position IDs to distinguish the left and right channels.

Temporal coordinates for audio and video are aligned with actual playback times. At 40 latent vectors per second per channel, each audio latent corresponds to 25 milliseconds. The beginning of the user input audio defines the common temporal origin. Each audio token receives a temporal position equal to the offset of its segment start from this origin, divided by 25 milliseconds. Video temporal positions use the corresponding playback times with the same origin and temporal unit. This accounts for the nonuniform temporal mapping from 17 source frames to five video latent frames.

\FloatBarrier
\section{Training Strategy}
\label{sec:training}
\begin{figure}[t]
  \centering
  \includegraphics[width=\linewidth]{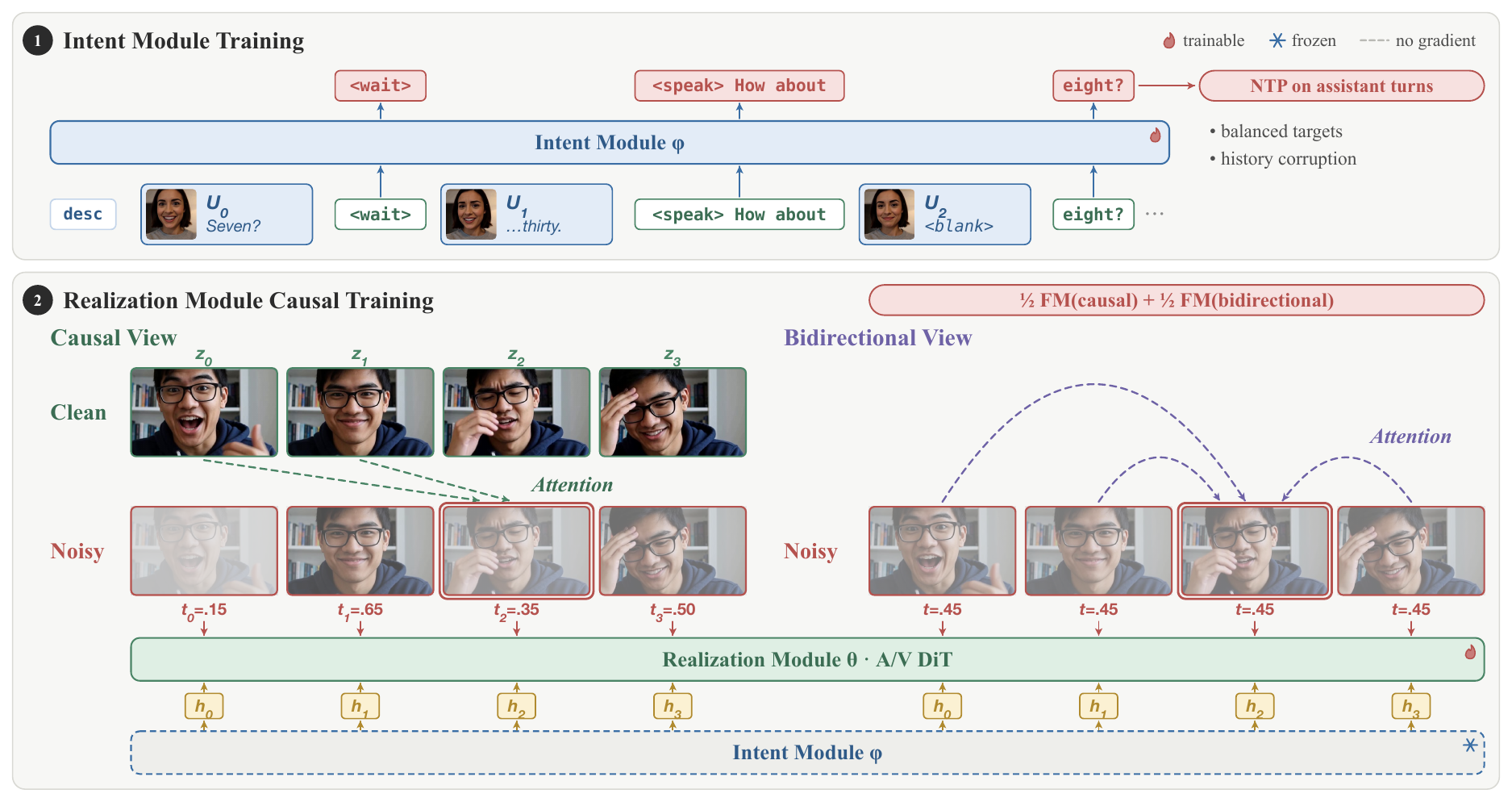}
  \caption{Overview of Stage 1 (Intent Training) and Stage 2 (Realization Training).}
  \label{fig:stage12}
\end{figure}

Training proceeds in three stages. Stage 1 trains the Intent Module for causal participation and text prediction through supervised fine-tuning. Stage 2 trains the Realization Module for causal joint audiovisual generation, conditioned on recorded media history and intent features extracted from recorded assistant text by the frozen Intent Module (\figref{fig:stage12}). Stage 3 distills four-step generation through distribution matching, with dual rollout using predicted text and generated audiovisual history to match inference-time conditioning.

\subsection{Stage 1: Intent Training}
\label{sec:intent-training}

We adapt the multimodal Intent Module with causal supervised fine-tuning on chunk-aligned conversations. At the start of chunk $k$, the model predicts the assistant output text $y_k$ from the global instruction $d$, the user video $u_v$ and ASR text $u_t$ available through chunk $k-1$, and earlier assistant language $y_{<k}$. The decision context excludes assistant video. In the training data, user tokens become available at the ends of their aligned spans, while assistant text tokens are assigned to the chunk containing their speech onset. The \code{<speak>} action introduces new text, including continuations of an earlier turn. The \code{<wait>} action adds no new text and can coexist with ongoing playback or non-verbal behavior. \figref{fig:intent-chunk-alignment} illustrates this shift, and \secref{sec:observation-adapter} details the input protocol.

\paragraph{Causal chunk supervision.} Let $y_k$ be the token sequence of the assistant output in chunk $k$, and let $C^{\mathrm I}_k=(d,u_{v,<k},u_{t,<k},y_{<k})$ denote the causal context of the Intent Module. We minimize the weighted autoregressive objective
\begin{equation}
 \mathcal L_{\mathrm{intent}}
 =-\mathbb E\!\left[
 \frac{\sum_{k,j}m_{k,j}w_{k,j}\log p_\phi\!\left(y_{k,j}\mid C^{\mathrm I}_k,y_{k,<j}\right)}
 {\max\!\left(1,\sum_{k,j}m_{k,j}w_{k,j}\right)}
 \right],
 \label{eq:planner-objective}
\end{equation}
where $y_{k,j}$ is the $j$-th action, content, or structural token, $w_{k,j}$ is its category-dependent weight, and $m_{k,j}$ is one for supervised tokens and zero otherwise. The denominator sums the effective weights within each sampled conversation, with a minimum of one. The initial assistant output $y_0$ is a fixed \code{<wait>} used as context and excluded from the loss.

\paragraph{Action-aware rebalancing.} Most chunks introduce no new assistant text, so unweighted training can favor waiting. We assign supervision categories offline from recorded assistant events and from whether the user is speaking at the chunk start. This speaking status distinguishes idle waits from waits during user speech and ordinary openings from interruption openings. The assistant annotation distinguishes openings from continuations, and an annotated question in the preceding user turn provides a separate opening tag. We use these labels to subsample targets and weight the loss, with separate action and content weights and adjustments for opening categories. Subsampled targets remain in the conversation history. The model receives only the causal context described above. Validation retains the full set of eligible targets and reports separate token-level metrics for actions, speech, and waits.

\paragraph{Training under imperfect assistant history.} To expose the model to imperfect histories, we truncate or repeat text in selected continuation chunks, mask their losses, and retain supervision on later valid targets. For missed-opening recovery, we replace selected ordinary openings with \code{<wait>} in the history and mask their losses. In the next chunk of the same assistant turn, only \code{<speak>} contributes to the loss. 

\paragraph{Preserving visual and linguistic grounding.} The training mixture contains examples whose correct answer depends on the user video as well as examples solvable from text alone. We identify visually grounded assistant turns through data inspection and upweight their content-token loss, while retaining the same causal visual input path. Prompt augmentation varies the format, ordering, and completeness of the static assistant/scene description without inventing new facts. Together with source-grouped train/validation splits, these choices discourage the Intent Module from relying on a fixed prompt template or on text-only shortcuts. Optional emotion labels are kept as separately weighted attachments and are admitted only at their annotated causal boundary. We use low-rank adaptation to fine-tune the Intent Module. After this stage, the frozen Intent Module supplies hidden states from its forward passes to condition media training.

\subsection{Stage 2: Realization Training}
\label{sec:mixed_training}

Stage 2 trains the Realization Module for causal joint audiovisual generation while keeping the Intent Module frozen. Training follows the conditional structure in \eqref{eq:realization-step}, learning to generate audio and video latents from the current intent representation and available interaction context. The sequence layout and attention mask encode the dependencies among modalities within each chunk and restrict access to preceding interaction history across chunks. We describe this multimodal sequence construction, the joint flow-matching formulation, and the optimization strategy below.

\begin{figure}[t]
  \centering
  \includegraphics[width=\textwidth]{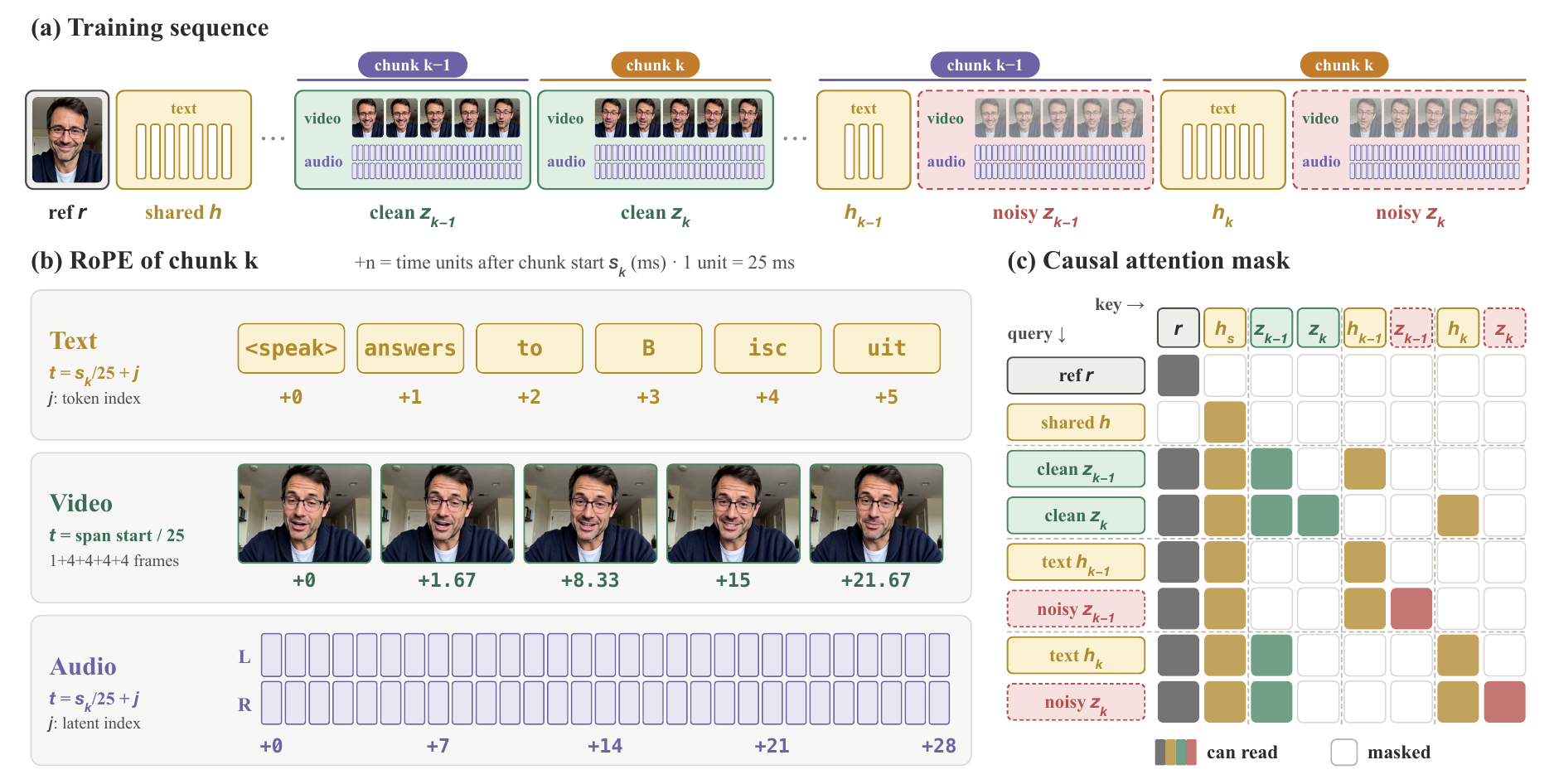}
  \caption{\textbf{Training sequence, positional encoding, and attention mask of the Realization Module.} (a) The training context $C^{\mathrm R}$ contains the reference, the shared global instruction, clean media of all chunks, and, for each chunk, the Intent Module hidden states followed by the noisy media. (b) Temporal RoPE positions in chunk $k$ follow playback time in 25\,ms units from the chunk start $s_k$: text tokens and audio latents advance by one unit per index, and each video latent takes the start time of its frame span. (c) Causal attention mask. The noisy media of chunk $k$ read the reference, the shared instruction, clean media of earlier chunks, the hidden states of chunk $k$, and themselves, but not the clean media of chunk $k$ or any later chunk.}
  \label{fig:attention-mask}
\end{figure}

\paragraph{Chunk structure and sequence design.} Each segment of streaming generation spans 17 video frames at 24 fps, or $17/24\approx0.708$ seconds. The chunk $\mathbf{c}_k$ contains the assistant action for that chunk, represented by the text embedding $x_k^{t}$ obtained from the Intent Module hidden state $h_k$, together with the corresponding noised assistant video and audio embeddings:
\begin{equation}
\mathbf{c}_k = (x_k^{t}, \tilde{x}_k^{v}, \tilde{x}_k^{a}),
\end{equation}
where $\tilde{x}_k^{v}$ and $\tilde{x}_k^{a}$ are the clean media embeddings $x_k^{v}$ and $x_k^{a}$ noised to level $\tau_k$. To align with the timelines used by the audio and video codecs, the number of media embeddings in each chunk corresponds to 5 latent video representations and $2\times29$ latent audio representations.

During training, we also insert clean audio and video embeddings into the sequence. A carefully designed attention mask prevents noisy media embeddings in each generation chunk from accessing their own clean signals. For each training example, the sequence contains, in order, the assistant reference-media embeddings, the global instruction embeddings, the clean media embeddings for the generation chunks, and the generation chunks themselves:
\begin{equation}
C^{\mathrm R} = (x^{r}, x^{d}, x_0^{v}, x_0^{a}, \ldots, x_K^{v}, x_K^{a}, \mathbf{c}_0, \ldots, \mathbf{c}_K),
\end{equation}
where $x^{r}$ embeds the reference image $I_{\mathrm{ref}}$, $x^{d}$ embeds the global instruction $d$, and $K+1$ is the number of chunks in the sample.

To fit within the context length of the initialized model, we randomly sample a window of consecutive chunks from each training conversation. Because temporal position IDs align with actual video playback times, positional encoding remains closely aligned between training and inference even when the sampled chunk sequence does not start at the beginning of a conversation.

\paragraph{Attention mask design.} Each noisy media embedding can attend to the reference media, global description, historical assistant text, and clean historical media. Clean media related to the current chunk and all content in future chunks are invisible. Noisy media embeddings within the current chunk also have internal content boundaries: video embeddings correspond in order to 5 video-encoder latent representations, while audio embeddings are divided into left and right channels, each corresponding to 25 milliseconds of audio. Nevertheless, all these embeddings are mutually visible within the chunk. The detailed visibility relationships are illustrated in \figref{fig:attention-mask}.

\paragraph{Flow-matching velocity prediction.} Flow matching~\citep{lipman2023flowmatching} supervises the joint media velocity from random noise to clean latent states, with clean time $\tau=1$ and noise time $\tau=0$. For clean media latents $z\in\{z^{v},z^{a}\}$ and Gaussian noise $\epsilon$, we have:
\begin{equation}
\begin{aligned}
\tilde{z}_{\tau} &= (1-\tau)\epsilon+\tau z
&\quad& \text{noised input}, \\
v^\star &= z-\epsilon
&& \text{target velocity}.
\end{aligned}
\label{eq:flow-foundation}
\end{equation}

\subsection{Stage 3: Distribution Matching Distillation}
\label{sec:dmd}

\paragraph{DMD supervision.}
DMD~\citep{yin2024dmd} distills the causal model into a four-step chunk-causal generator $G_\theta$ by matching the distribution of its own samples to the data distribution, instead of enforcing agreement between neighboring points on a teacher trajectory as CD does. 
Two score models provide this supervision: a frozen real-score model $\mu_{\mathrm{real}}$ approximates the conditional data distribution, and a trainable fake-score model $\mu^{\psi}_{\mathrm{fake}}$ tracks the evolving generator distribution.
The generator synthesizes a latent window $z_{\mathrm{gen}}$ causally, whereas both score models evaluate the completed window bidirectionally. In our flow convention ($\tau=0$ noise, $\tau=1$ data), each conversation draws one noise level $\tau$ from a shifted discrete noise schedule, the window is corrupted once as $\tilde z_{\mathrm{gen}}=\tau z_{\mathrm{gen}}+(1-\tau)\epsilon$, and the two models return clean estimates $\hat z_{\mathrm{real}}=\mu_{\mathrm{real}}(\tilde z_{\mathrm{gen}},\tau)$ and $\hat z_{\mathrm{fake}}=\mu^{\psi}_{\mathrm{fake}}(\tilde z_{\mathrm{gen}},\tau)$ under this shared corruption. Since a clean estimate $\hat z_\ast$ determines the score $s_\ast=-(\tilde z_{\mathrm{gen}}-\tau\hat z_\ast)/(1-\tau)^2$, the DMD gradient of the reverse KL divergence reduces to the difference of the two estimates,
\begin{equation}
\nabla_\theta D_{\mathrm{KL}}\!\left(p_\theta\,\|\,p_{\mathrm{data}}\right)
\approx\mathbb E_{\tau,\epsilon}\!\left[\left(s_{\mathrm{fake}}-s_{\mathrm{real}}\right)^{\!\top}
  \frac{\partial z_{\mathrm{gen}}}{\partial\theta}\right],
\qquad
s_{\mathrm{fake}}-s_{\mathrm{real}}=\frac{\tau}{(1-\tau)^{2}}\left(\hat z_{\mathrm{fake}}-\hat z_{\mathrm{real}}\right).
\label{eq:dmd-gradient}
\end{equation}
We replace the time-dependent factor by a sample-adaptive scale and realize the update as regression toward a detached target; for one conversation and one modality,
\begin{equation}
\kappa=\max\!\left(\left\langle|z_{\mathrm{gen}}-\hat z_{\mathrm{real}}|\right\rangle_w,\ \varepsilon_0\right),
\qquad
\Delta=\frac{\hat z_{\mathrm{fake}}-\hat z_{\mathrm{real}}}{\kappa},
\qquad
z_{\mathrm{DMD}}=\operatorname{sg}\!\left[z_{\mathrm{gen}}-\Delta\right],
\label{eq:dmd_target}
\end{equation}
where $\operatorname{sg}$ denotes stop-gradient, $\langle\cdot\rangle_w$ averages over feature dimensions and then over the valid target rows of that conversation and modality, and the floor $\varepsilon_0$ prevents division by a vanishing scale. Assembling the targets across conversations yields the generator loss
\begin{equation}
\mathcal L_{\mathrm{DMD}}=\frac12\sum_{m\in\{v,a\}}\lambda_m\,
  \operatorname{WMSE}_{w_m}\!\left(z^{m}_{\mathrm{gen}},z^{m}_{\mathrm{DMD}}\right),
\label{eq:dmd-loss}
\end{equation}
whose WMSE reduction uses the global valid-target normalization of \secref{sec:mixed_training}. Its gradient with respect to $z_{\mathrm{gen}}$ is proportional to $\Delta$, so the generator is moved along $(\hat z_{\mathrm{real}}-\hat z_{\mathrm{fake}})/\kappa$ through its differentiable outputs.

Because the fake model must follow the generator, it is fitted on detached generated samples by flow matching, $\mathcal L^{\mathrm{fake}}_{\mathrm{FM}}=\mathbb E\big[\omega(\tau)\, \|v_\psi(\tilde z_{\mathrm{gen}},\tau)-(z_{\mathrm{gen}}-\epsilon)\|^{2}\big]$, where $\omega(\tau)$ is a noise-level-dependent weight. The fake model is updated more frequently than the generator, with several fake updates between consecutive generator updates. Each update draws its own window and rollout, so the fake model is always fitted to samples from the current generator.

\label{sec:gan-refinement}
An optional GAN loss~\citep{yin2024dmd2} adds direct feedback from training media. A lightweight head on middle-block features of the fake-score backbone pools the valid audio and video tokens of each conversation through a learned attention query into a single logit $D(\cdot)$. Its input physically excludes text, Intent Module, reference, and clean-context rows, so the critic judges joint audio--visual realism without requiring generated media to reproduce the recorded transcript. Writing $\tilde z_{\mathrm{data}}$ and $\tilde z_{\mathrm{gen}}$ for training-media and generated latents noised to the same level with independent noise, the two trainable roles minimize
\begin{equation}
\begin{aligned}
\mathcal L_{G}&=\mathcal L_{\mathrm{DMD}}
  +\lambda_G\,\mathbb E\!\left[\operatorname{softplus}\!\left(-D(\tilde z_{\mathrm{gen}})\right)\right],\\
\mathcal L_{F}&=\mathcal L^{\mathrm{fake}}_{\mathrm{FM}}
  +\lambda_D\,\mathbb E\!\left[\operatorname{softplus}\!\left(-D(\tilde z_{\mathrm{data}})\right)
  +\operatorname{softplus}\!\left(D(\tilde z_{\mathrm{gen}})\right)\right],
\end{aligned}
\label{eq:combined-distillation}
\end{equation}
where the fake backbone and head are frozen but differentiable in the generator update, and generated media are detached in the critic update.

\begin{figure}[t]
  \centering
  \includegraphics[width=\linewidth]{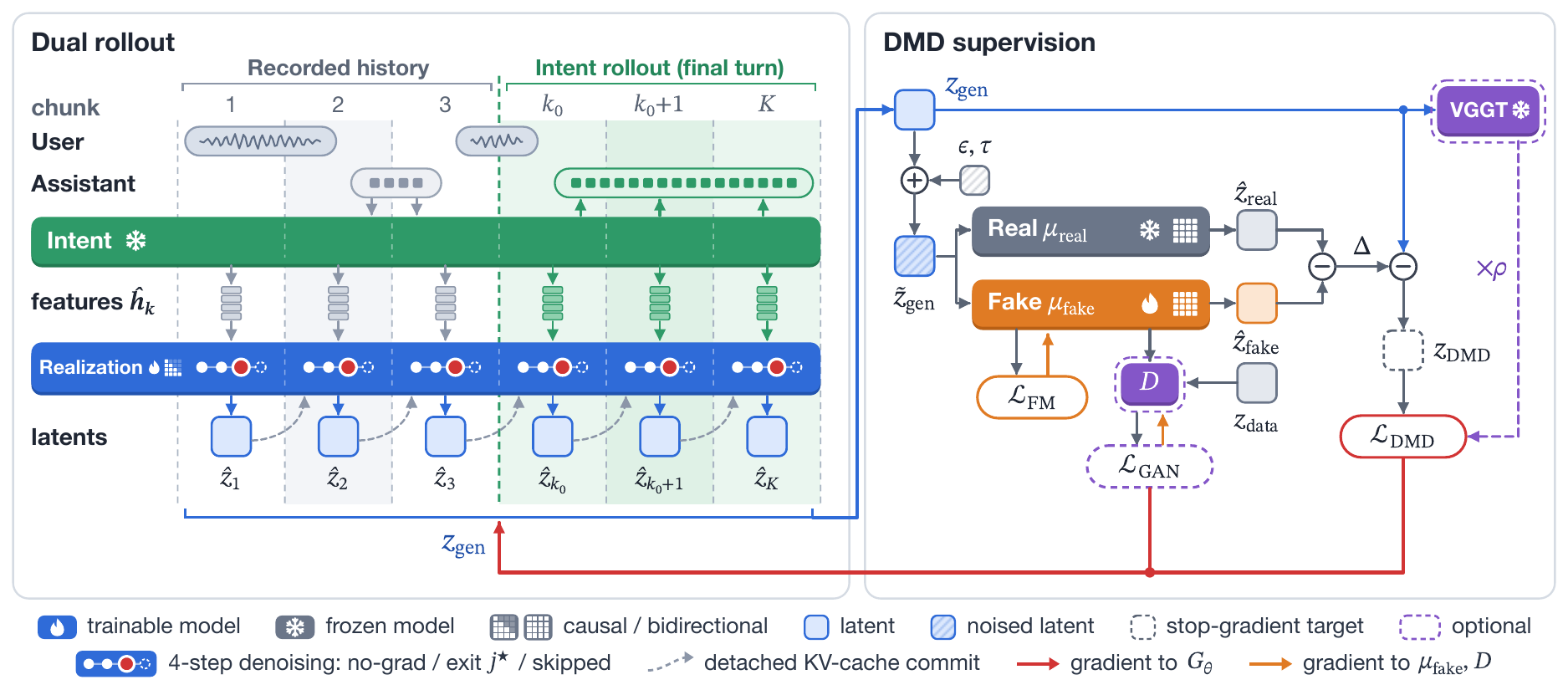}
    \caption{\textbf{DMD training with dual rollout.} The frozen Intent Module replays recorded assistant text and generates the final assistant turn from chunk $k_0$ onward. Its hidden states $\hat h_k$ condition the Realization Module $G_\theta$, which generates the latent window $z_{\mathrm{gen}}$ chunk by chunk. The frozen real score model $\mu_{\mathrm{real}}$ and the trainable fake score model $\mu_{\mathrm{fake}}$ evaluate the corrupted latent window to construct the target $z_{\mathrm{DMD}}$. Red arrows indicate gradients to $G_\theta$. Orange arrows indicate gradients to $\mu_{\mathrm{fake}}$ and $D$. Dashed outlines indicate optional components.}
  \label{fig:dmd-training}
\end{figure}

\paragraph{Dual rollout in DMD training.}
\label{sec:dual-rollout}
Dual rollout aligns DMD training with the language and media context encountered during inference (\figref{fig:dmd-training}). Before DMD, the Realization Module conditions on features extracted from layer 50 of the Intent Module using recorded assistant text, together with recorded media prefixes. At inference, the Intent Module determines response wording, length, and speaking timing, while the Realization Module continues from its own audiovisual outputs. Each DMD sample therefore undergoes a language rollout by the frozen Intent Module, followed by a media rollout by the Realization Module.

The frozen Intent Module rolls out only the final assistant turn of each training window. Windows may start at any earlier chunk and end at a natural assistant turn boundary rounded up to a codec chunk. Candidates are bucketed by reserved token length, with bucket weights proportional to squared length. Let $k_0$ be the first chunk within the window containing the final turn. Using only history within the window (\secref{sec:foundations}), the Intent Module replays recorded text before $k_0$ and greedily predicts participation actions and text thereafter. Recorded tokens of the final turn and all later text are removed, while any recorded prefix from an earlier assistant turn sharing the boundary chunk is retained. Media rollout follows Self Forcing~\citep{yin2025selfforcing}:
\begin{equation}
\begin{aligned}
\widehat h_k&=
\begin{cases}
\operatorname{Intent}_\phi\!\left(C^{\mathrm I}_k;\,y_k\right), & k<k_0\quad\text{(recorded history)},\\[2pt]
\operatorname{Intent}_\phi\!\left(C^{\mathrm I}_k\right), & k\ge k_0\quad\text{(intent rollout)},
\end{cases}\\
\widehat z_k&=G^{(j^\star)}_\theta\!\left(\epsilon_k;\,x^{r},x^{d},\widehat h_k,\operatorname{sg}[\widehat z_{<k}]\right),
\qquad j^\star\sim\mathcal U\{0,\dots,N-1\},
\end{aligned}
\label{eq:dual-rollout}
\end{equation}
Here, $C^\phi_k$ contains recorded user inputs and earlier Intent Module outputs within the window, $x_d^t$ embeds the global instructions, and $\widehat h_k$ are layer 50 states captured during the actual forward passes without an additional pass over generated text. With $N=4$, $G_\theta^{(j)}$ performs $j$ steps without gradient tracking followed by one differentiable step whose clean estimate becomes the chunk output. The exit $j^\star$ is shared across the trajectory. Each detached clean output is processed by a clean forward at $\tau=1$ to update the KV cache. The assembled window $z_{\mathrm{gen}}=(\widehat z_k)_k$ supplies the targets and losses in \eqref{eq:dmd_target} and \eqref{eq:combined-distillation}. The recorded turn boundary defines the training window without constraining the length or content of the generated response.

\paragraph{Auxiliary camera-rotation-aware weighting.}
\label{sec:quality_weighting}
DMD can introduce camera-rotation drift that distribution matching does not explicitly penalize. We therefore reweight the video DMD loss using rotations $R_i$ estimated by frozen VGGT~\citep{wang2025vggt} from decoded rollout frames. Let $\alpha_i$ denote the geodesic angle between $R_0$ and $R_i$, and $\delta_i$ that between consecutive rotations. Sustained deviation and accumulated rotation define
\begin{equation}
\begin{aligned}
E&=\frac{1}{T_0}\int\!\left(\frac{[\alpha(s)-\alpha_{\mathrm{dz}}]_+}{\alpha_{\mathrm s}}\right)^{\!2}ds
  +\left(\frac{1}{\alpha_{\mathrm p}}\sum_i\delta_i\,\frac{g_{i-1}+g_i}{2}\right)^{\!2},
\qquad
g_i=\operatorname{clip}\!\left(\frac{\alpha_i-\alpha_{\mathrm{dz}}}{\alpha_{\mathrm g}},0,1\right),\\
\rho&=1+\lambda_{\mathrm{rot}}\left(\max\!\left(\rho_{\max}e^{-E},\,\rho_{\min}\right)-1\right),
\end{aligned}
\label{eq:vggt-weight}
\end{equation}
Here, $s$ is playback time, $\alpha_{\mathrm{dz}}$ defines a dead zone, and $\alpha_{\mathrm s}$, $\alpha_{\mathrm g}$, and $\alpha_{\mathrm p}$ scale deviation, gating, and rotational path length, respectively. The constant $T_0$ normalizes duration, with weight bounds $0<\rho_{\min}<1<\rho_{\max}$ and strength $\lambda_{\mathrm{rot}}\in[0,1]$. The detached weight $\rho$ scales only the video term in \eqref{eq:dmd-loss}. At $\lambda_{\mathrm{rot}}=1$, stable windows receive weights up to $\rho_{\max}$, while strongly rotating windows receive $\rho_{\min}$. The metric uses rotation alone, and opposite rotations do not cancel. This weighting rebalances DMD updates without propagating gradients through VGGT or the VAE. Intentional viewpoint changes and unreliable rotation estimates can also affect the weights.

\Needspace{7\baselineskip}
\section{Data Construction}
\label{sec:data}
\suppressfloats[t]

Training \methodname requires temporally aligned user inputs and assistant audiovisual behavior. We combine synthetic and recorded interactions into a unified conversation timeline, preserving natural timing and controlled events. This provides causal supervision for Intent and audiovisual targets for Realization, with additional routes for missing-view reconstruction and recorded-video extensions.

\subsection{Synthetic Data}
\label{sec:data-sources}
\label{sec:synthetic-data}

\begin{figure}[t]
  \centering
  \includegraphics[width=\linewidth]{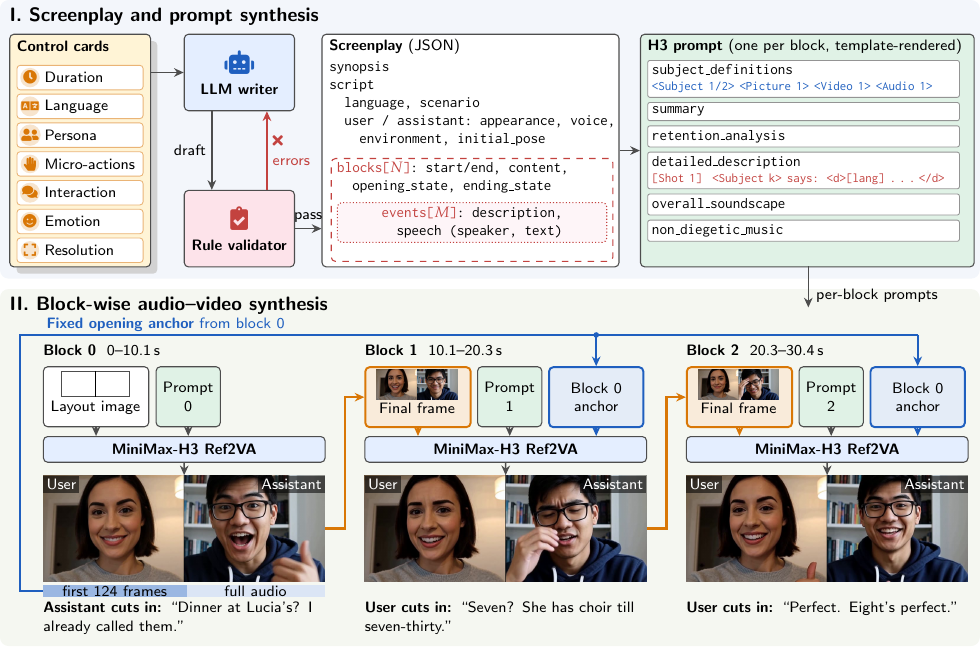}
  \caption{\textbf{Screenplay-driven dual-view generation.}
  Stage~I turns sampled control cards into a validated structured screenplay and renders it, without another LLM call, into one t2v prompt per block. Stage~II generates one split-screen audio-video block by block. Each later block starts from the previous block's final frame (orange) and is anchored to the same opening segment of block~0 (blue), so identity and voice references are never re-chained through earlier continuations. Frames and dialogue in this figure come from one generated 30\,s sample.}
  \label{fig:dual-view-pipeline}
\end{figure}

\paragraph{Pipeline overview.}
This route generates a two-person video call in which both participants are rendered together. \figref{fig:dual-view-pipeline} shows its two stages. \textbf{Stage~I} is text only. Control cards are sampled, a LLM writes a structured screenplay under these cards, rule checks reject invalid screenplays, and a template converts each screenplay block into a text prompt~\citep{h3modelcard}. \textbf{Stage~II} feeds these prompts to a t2v model, which generates one split-screen audio-video stream block by block. The left panel is always the user and the right panel the assistant. Generating both panels and the shared audio on a unified timeline preserves the timing of dialogue, reactions, overlaps, and interruptions.

\paragraph{Control cards.}
Without explicit constraints, LLM-generated conversations tend to repeat similar topics, relationships, and turn-taking patterns. To improve diversity, we use control cards that define attributes with predefined candidate values. For each screenplay, these values are sampled uniformly and included as explicit requirements in the generation prompt.
\begin{itemize}[leftmargin=*, itemsep=0pt, topsep=2pt, parsep=0pt]
  \item duration: \{30\,s, 45\,s, 60\,s\}
  \item language: \{Chinese, English\}
  \item persona: \{none, relationship, occupation\}, \textit{e.g.}, relationship $\in$ \{``parent-child'', ``colleagues'', ``doctor-patient'', \dots\} (12 values), occupation $\in$ \{``firefighter'', ``nurse'', ``chef'', \dots\} (11 values, per participant)
  \item micro-actions: \{off, on\}, where ``on'' suggests 24 small reactions, \textit{e.g.}, \{``nod'', ``raise an eyebrow'', ``look away briefly'', \dots\}
  \item interaction: \{``everyday exchange'', ``interruption'', ``follow-up / debate'', ``proactive inquiry''\}
  \item emotion: fixed to ``follow the topic and the conversation''
  \item resolution: \{$1536\times576$, $1152\times768$\}, used only by video synthesis
\end{itemize}
Each \texttt{interaction} value carries a written specification. For example, ``interruption'' requires one speaker to cut in before the other finishes, a brief overlap of voices, and the interrupted speaker yielding the turn.

\begin{figure}[t]
  \centering
  \includegraphics[width=\linewidth]{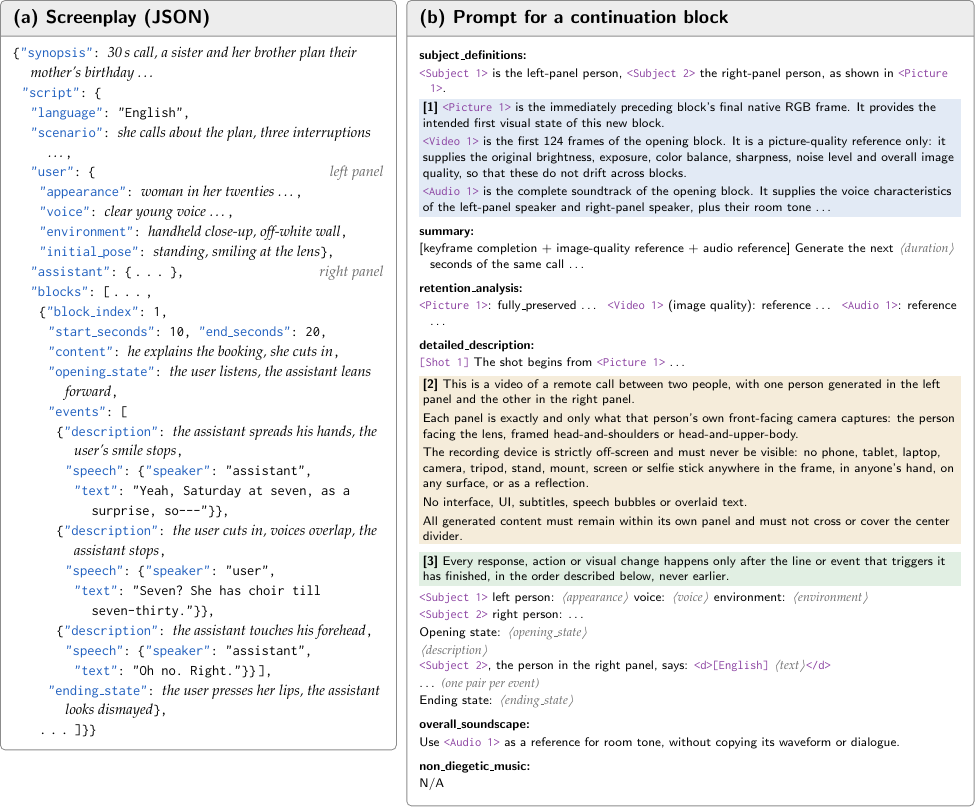}
  \caption{\textbf{Screenplay and t2v prompt.}
  (a) The screenplay written by the LLM for the sample in \figref{fig:dual-view-pipeline}. (b) The t2v prompt rendered from it for a continuation block. Shaded sentences are fixed text shared by all samples, grouped as [1]-[3] in the main text. \textcolor{gray}{$\langle\cdot\rangle$} marks screenplay fields copied into the prompt, and the remaining lines are abridged template text.}
  \label{fig:screenplay-prompt}
\end{figure}

\paragraph{Screenplay and t2v prompt synthesis.}
The sampled cards are inserted into a fixed template, and the LLM writes a synopsis followed by a JSON screenplay, shown in \figref{fig:screenplay-prompt} (a). The screenplay describes each participant once (appearance, voice, environment, initial\_pose) and then splits the call into blocks of 5-15\,s at natural turn boundaries. Each block is an ordered list of events. An event pairs a description of expressions, gestures, and the listener's reaction with at most one speech entry, so nonverbal reactions need no invented dialogue. A rule-based validator checks the structure, block timing, language, and the absence of capture-device words such as ``phone'' in the narration, and returns errors to the LLM for up to 3 attempts.

The validated screenplay is converted into one t2v prompt per block by a template, without further LLM calls (\figref{fig:screenplay-prompt}(b)). Screenplay fields fill the slots verbatim. The shaded sentences are fixed for all samples and fall into three groups:
\begin{itemize}[leftmargin=*, itemsep=1pt, topsep=2pt, parsep=0pt]
  \item \textbf{Reference roles.} Each input is assigned one job: the previous final frame is the starting state, the opening video only sets picture quality, and the opening audio only sets voices and room tone. Without these roles the model tends to replay the reference video or repeat its dialogue.
  \item \textbf{Picture content.} Each panel is the participant's own front camera, which rules out third-person, mirror, and behind-the-back views. Recording devices, UI elements, and subtitles, which otherwise appear frequently in generated calls, are excluded, and content may not cross the center divider, so the two crops stay clean.
  \item \textbf{Causal order.} A reaction may start only after its trigger has finished, so the assistant view never responds before the user acts.
\end{itemize}

\begin{figure}[t]
  \centering
  \includegraphics[width=\linewidth]{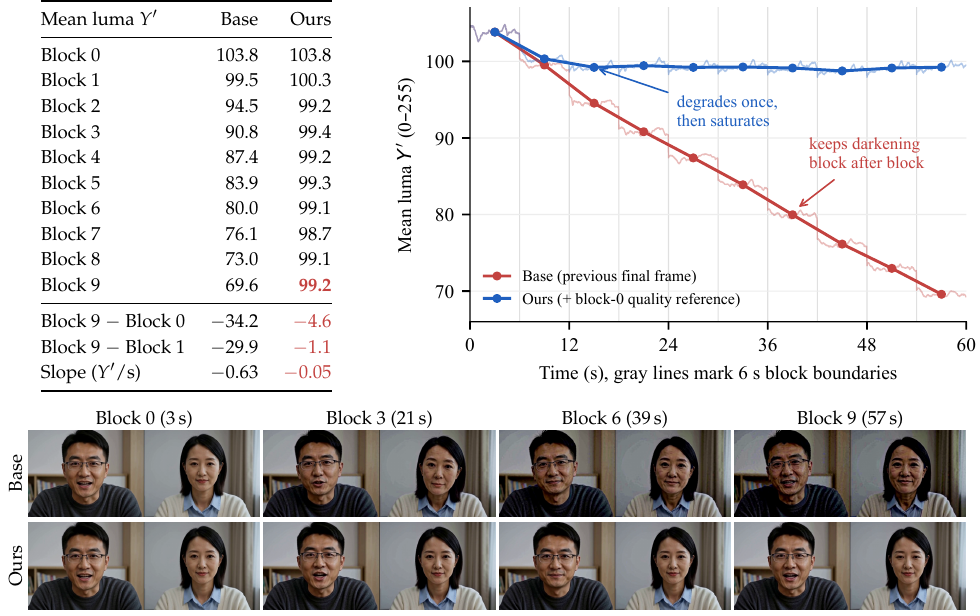}
  \caption{\textbf{Error accumulation over a 60\,s, ten-block rollout.}
  Top: mean luma $Y'$ (0--255) of the full split-screen frame per block
  (table) and per frame (plot). Bottom: the split-screen frame at the
  middle of blocks 0, 3, 6, and 9. Base conditions each block only on the
  previous final frame, and Ours adds the fixed block-0 picture-quality
  reference. Both runs share the same block~0 and screenplay.}
  \label{fig:brightness-drift}
\end{figure}

\paragraph{Video synthesis.} 
As shown in \figref{fig:dual-view-pipeline}, Block~0 receives its prompt and a white layout image whose two panels are separated by a thin black divider. The layout image fixes only the split-screen geometry, and the model creates both participants and backgrounds from the prompt. 
\begin{itemize}[leftmargin=*, itemsep=1pt, topsep=2pt, parsep=0pt]
  \item \textbf{Previous final frame} (orange). The last frame of block $i-1$ is temporally aligned with the first target frame, preserving pose, expression, and framing across blocks.
  \item \textbf{Opening video} (blue). The first 124 frames of block~0 serve as a picture-quality reference for brightness, exposure, color sharpness, and noise.
  \item \textbf{Opening audio} (blue). The full block~0 audio anchors both voices and room tone.
\end{itemize}
We preserve generation conditions in their original representations, using VAE latents for the initial video and raw RGB for the final frame. The generated video are then concatenated and split into user and assistant views, both sharing the same synchronized audio track.

\paragraph{Error accumulation.} 
The baseline data synthesis method conditions each block on the final frame of the preceding block, causing visual errors to accumulate over time. Our synthesis method mitigates this drift by using the opening video (block~0) as a fixed reference throughout generation. \figref{fig:brightness-drift} compares the two synthesis strategies over a 60\,s sequence of ten 6\,s blocks. The baseline loses 34.2 luma levels by block~9, with increasingly exaggerated facial textures. Our method stabilizes at approximately 99 luma after a slight initial decline, maintaining consistent facial appearance through block~9.

\subsection{Recorded Data}
\label{sec:recorded-data}

\begin{figure}[t]
  \centering
  \includegraphics[width=\linewidth]{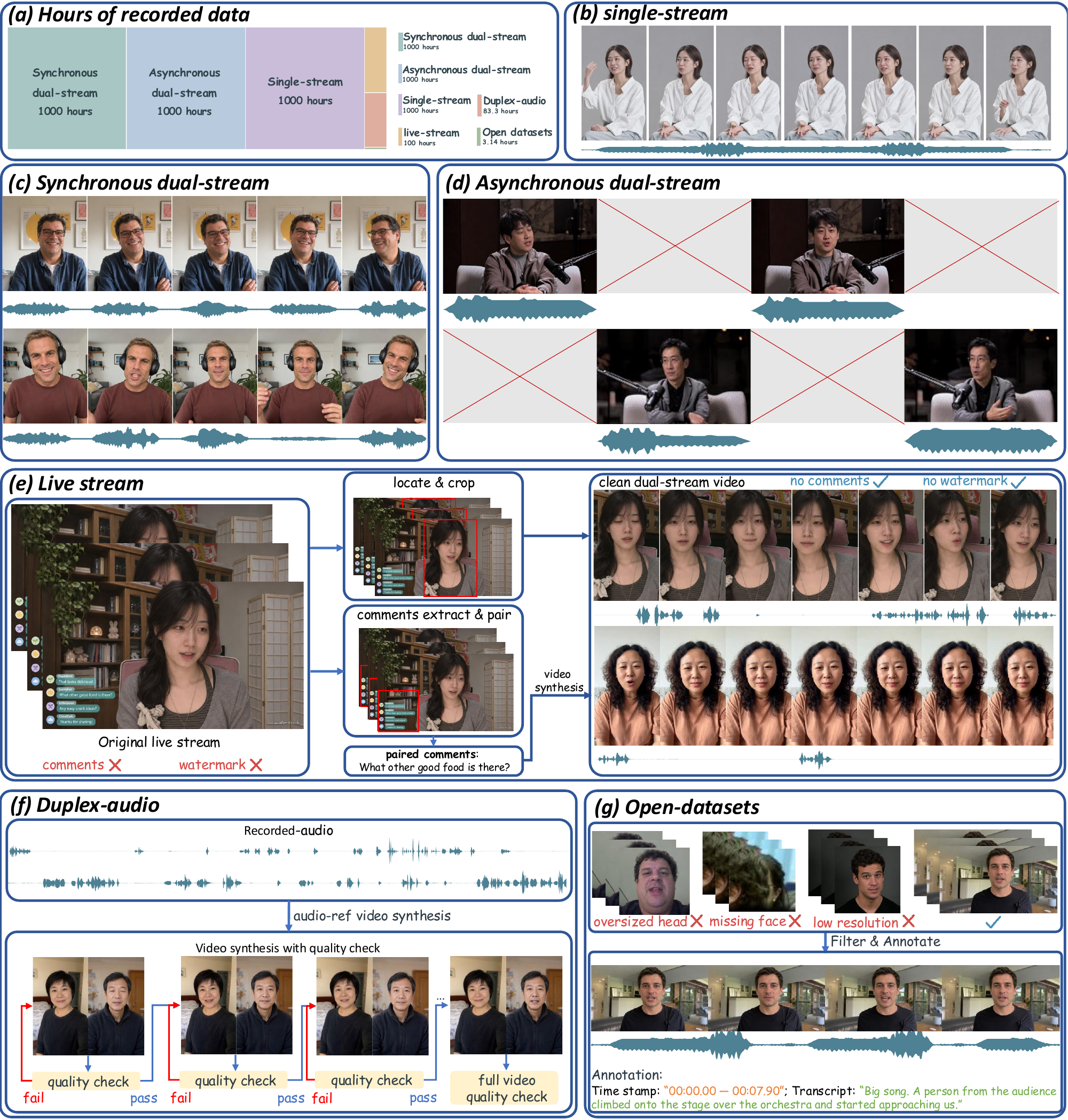}
  \caption{\textbf{Recorded data sources, interaction formats}
  The six categories describe the recorded sources, with selected categories requiring further processing or synthesis to construct training examples.
  }
  \label{fig:recorded-data-overview}
\end{figure}

As shown in \figref{fig:recorded-data-overview}, we organize recorded data into six categories by source and interaction format: single-stream video, synchronous dual-stream video, asynchronous dual-stream video, live streams with audience comments, duplex audio recordings and open datasets.
The following common recording standards apply across these categories.
Recordings feature real speakers and their original voices, with speaker identities, transcripts, and temporal mappings retained and unrelated audio excluded.
For video data, each delivered clip preserves a continuous, uncut half-body or full-body view with accurate audio--visual synchronization, at least 720p effective resolution per view, and no visual overlays.
Dual-stream recordings contain at least one complete conversational exchange.

\paragraph{Single-stream video.}
\figref{fig:recorded-data-overview}~(b) illustrates single-stream recordings, such as solo podcasts and single-person presentations.
The clips range from 10 to 90 seconds, preserving natural pauses.

\paragraph{Synchronous dual-stream video.}
Synchronous dual-stream video, illustrated in \figref{fig:recorded-data-overview}~(c), keeps both participants visible at the same time and preserves listening behavior, overlapping speech, and reactions.

\paragraph{Asynchronous dual-stream video.}
Asynchronous dual-stream video shows participants in separate clips that alternate with the speaking turns, as in the interview footage shown in \figref{fig:recorded-data-overview}~(d).

\paragraph{Live streams with audience comments.}
We align timestamped audience questions with presenter responses and reconstruct user videos, converting live streams into synchronized dual-stream conversations.

\paragraph{Duplex audio recordings.}
This category consists of full-duplex audio recordings of two-person dialogues, which drive the further video synthesis, as illustrated in \figref{fig:recorded-data-overview}~(f).

\paragraph{Open datasets.}
We collected single-stream talking videos from open-source datasets. As depicted in \figref{fig:recorded-data-overview}~(g), we filtered the material according to the common recording standards (e.g., resolution requirements) and then applied face detection, video understanding, voice activity detection (VAD), automatic speech recognition (ASR), and language identification to guide further filtering and annotation.

\subsection{Data Processing}
\label{sec:data-admission}

\subsubsection{Synthetic Data Quality Control}
Each newly generated video segment undergoes four automatic quality checks.
\paragraph{First-frame consistency.}
We compare the first frame with the conditioning image, requiring an SSIM of at least 0.90 and a perceptual-hash Hamming distance of at most 4. Conditioning uses the character image initially, then the preceding segment's processed final frame.

\paragraph{Unexpected audio.}
We compare the generated audio with the source audio using loudness normalization, waveform residuals, and VAD, rejecting segments with unexplained audio activity lasting at least 0.40 s.

\paragraph{Camera push-in.}
We track face scale at 4 FPS, using the first valid detection as the baseline. Rejection thresholds for sustained and abrupt enlargement are a smoothed scale of 1.30 times baseline for 0.75 s and a 1.22-fold jump within 0.50 s, respectively.

\paragraph{Persistent large-area visual changes.}
We sample frames at 4 FPS and measure LAB differences, changed-pixel proportions, and color/brightness histogram distances. Using these metrics, we reject abrupt changes that remain above thresholds relative to the pre-change frame, or cumulative drift from the initial frame exceeding both appearance and histogram thresholds.

\subsubsection{Live-Stream Data Processing}

\paragraph{Collection and initial screening.}
We collect archived live streams and deduplicate source records. Sampled frames are used to check the presence of real participants and visual quality, retaining only suitable material.

\paragraph{Annotation and filtering.}
We produce transcripts with word- or character-level timestamps by Qwen3-ASR-1.7B and Qwen3-ForcedAligner-0.6B~\citep{qwen2026asr}. And we also use Silero VAD~\citep{sileroteam2024vad} to identify speech and silence intervals and Qwen2.5-Omni-7B~\citep{xu2025qwen25omni} to review audio types. Speaker diarization and speaker-count checks combine FSMN-VAD~\citep{gao2023funasr} for speech detection, Paraformer ~\citep{gao2022paraformer}for transcription, and CAM++~\citep{wang2023campplus} for speaker embeddings followed by clustering. These checks exclude windows containing singing, whispered or breathy speech, or multiple speakers, producing timestamped candidate material for dialogue construction.

\paragraph{Dialogue construction and video preparation.}
To obtain the content and timing needed for user-side video synthesis, we match audience comments extracted from real chat logs to the presenter's responses. When such matching is not possible, we will generate counterpart utterances and schedule their interjections based on the presenter's original speech and available silence. After reviewing conversational coherence, completeness, and audio-visual correspondence, we extract continuous assistant-side video segments and crop them around the presenter to remove on-screen comments and watermarks. We then synthesize the corresponding user-side videos and pair the two views on a shared timeline. After a final quality check, we deliver clean dual-stream videos together with their transcripts, dialogue timing and source metadata.

\section{Streaming Pipeline and Acceleration}
\label{sec:runtime}

\suppressfloats[t]
In this section, we detail our streaming inference design. The system employs an orchestrated scheduling strategy to synchronize the intent module, the realization module and the VAE decoding, ensuring low-latency performance for interactive tasks.

\begin{figure}[t]
  \centering
  \includegraphics[width=\linewidth]{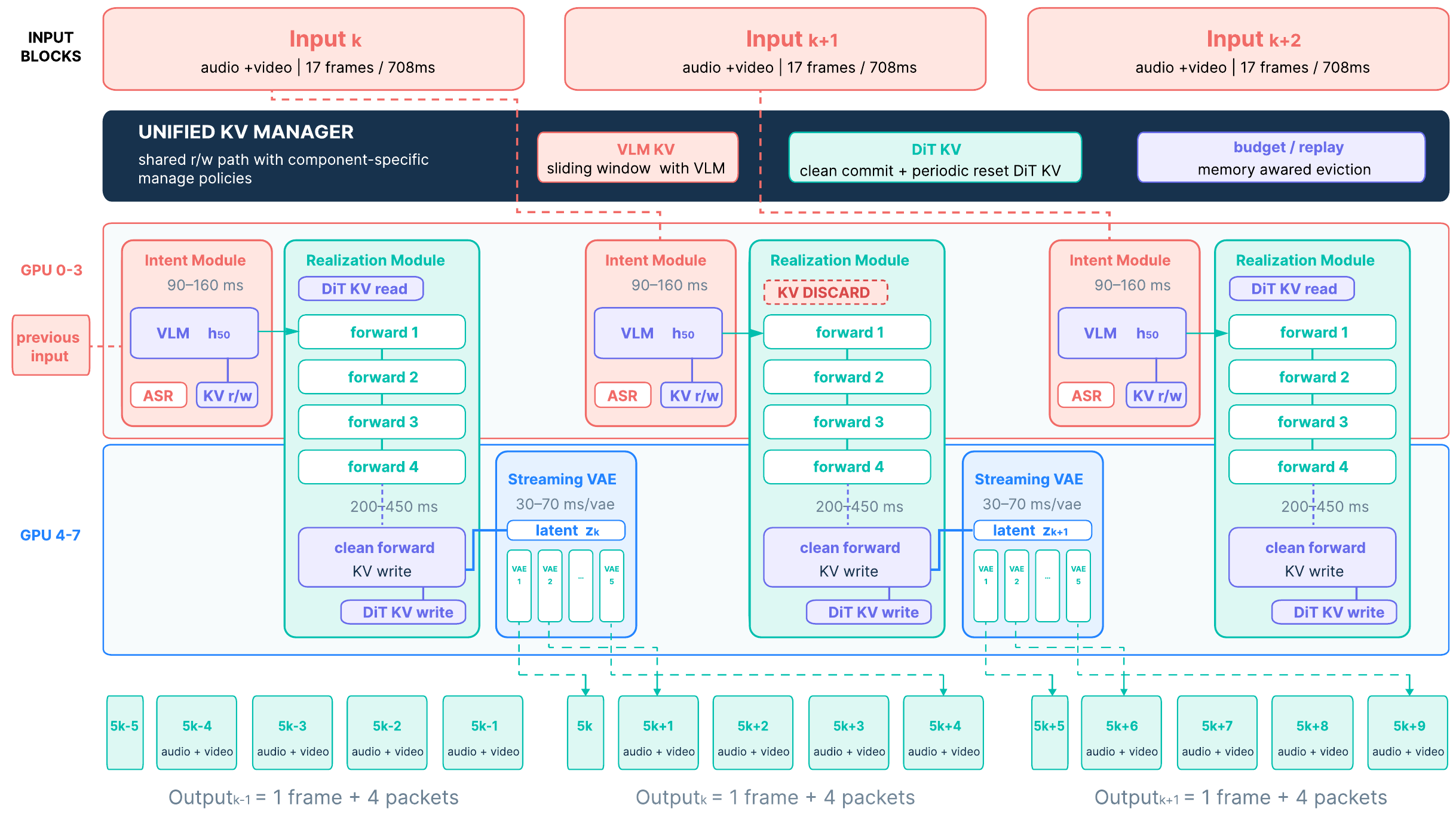}
  \caption{\textbf{Streaming inference on eight GPUs.}
  Time advances from left to right. The Intent Module runs its VLM with
  four-way tensor parallelism (TP4) on GPU~0-3, alongside auxiliary ASR. The Realization Module executes the DiT on GPU~0-7 using tensor or sequence parallelism (TP/SP), while the streaming VAE uses 4-way spatial tiling on GPU~4-7.
  The first visible execution of the Intent Module processes input $k$, which is omitted from the figure. The second execution processes input $k+1$. Hidden states $h_k$ from the Intent Module condition the Realization Module, which performs 4 denoising steps followed by a clean forward that commits the KV state.
  The unified KV manager uses a sliding window for Intent Module KV and periodic resets for Realization Module KV. Chunk $k+1$ skips history reads but retains the clean write. Decoding overlaps intent inference on separate GPUs before generation resumes. KV boxes mark manager operations.  
  }
  \label{fig:runtime}
\end{figure}

\subsection{Orchestrated Streaming Pipeline}
\label{sec:runtime-schedule}

The front-end device continuously captures user audio and video and forwards them for chunk-wise processing. As shown in \figref{fig:runtime}, each chunk passes through input acquisition, intent inference, joint latent generation, and streaming decoding, while input capture continues during response generation. The sequential stages of the pipeline can be summarized as:

\begin{itemize}[leftmargin=*, itemsep=1pt, topsep=2pt, parsep=0pt]
\item \textbf{Interaction sampling.} At 24 frames per second, each sampling window contains 17 video frames and the corresponding user audio, matching the video VAE grouping described in \secref{sec:vae}. The window duration is $t_{\mathrm{sample}} = 17/24\,\mathrm{s} \approx 708\,\mathrm{ms}$.
\item \textbf{Intent inference.} The Intent Module processes the available context, generates a participation and text output, and provides hidden states to the Realization Module. Depending on the input and context length, inference takes $t_{\mathrm{intent}} = 90\text{--}160\,\mathrm{ms}$.
\item \textbf{Joint latent generation.} Conditioned on the intent representation and media context, the Realization Module jointly generates audio and video latents. For video generation at 768$\times$576 resolution, this stage takes $t_{\mathrm{gen}} = 200\text{-}450\,\mathrm{ms}$, depending on the context length.
\item \textbf{Streaming decoding.} The generated audio and video latents are decoded incrementally for playback. Decoding a 4-frame group takes approximately $t_{\mathrm{decode}} = 30\,\mathrm{ms}$.
\item \textbf{Other overhead.} Data movement, transmission to the front-end device, and display processing contribute approximately $t_{\mathrm{overhead}} = 40\,\mathrm{ms}$.
\end{itemize}

The reported intent inference, latent generation, and overhead terms sum to
\begin{equation}
t_{\mathrm{intent}} + t_{\mathrm{gen}} + t_{\mathrm{overhead}}
\approx 330\text{--}650\,\mathrm{ms}.
\end{equation}
The delay to the first visible output additionally includes the initial decoding time and any remaining wait for the sampling window to close. An utterance or action ending within a sampling window can incur an additional sampling delay of up to approximately $708\,\mathrm{ms}$. During playback, we post-process the generated frames to maintain visual stability and smooth motion.

\subsection{Causal Latent Decoding}

The original VAE in~\citep{h3modelcard} decodes 5 latent frames into 17 video frames and requires 2 latent frames of lookahead. This dependency introduces additional waiting at chunk boundaries during streaming playback. We therefore adapt the decoding procedure to process available video latents incrementally.

After joint generation of a response chunk, its video latents are decoded one at a time. Each latent produces four video frames, followed by temporal alignment to match the original playback schedule and preserve the 17-frame chunk duration. This procedure allows playback to begin from the available chunk without waiting for additional latents from the next chunk.

At 24 frames per second, the shortest playback interval corresponds to the first latent in each group, which represents one video frame after temporal alignment. This interval is $1/24\,\mathrm{s} \approx 41.7\,\mathrm{ms}$. The measured per-latent decoding time is approximately $30\,\mathrm{ms}$, below this interval. The decoder can therefore keep pace with playback provided that generated latents are available on schedule.

\subsection{Inference Acceleration}
\label{sec:runtime-acceleration}

The runtime coordinates memory management, stage scheduling, and GPU placement to sustain continuous audiovisual generation. Each input chunk covers $17/24\,\mathrm{s}$ of captured audio and video. Sustained playback requires the pipeline to produce output chunks at least as quickly as they are consumed, while keeping memory usage bounded over long interactions. As shown in \figref{fig:runtime}, the system runs on a single server with eight GPUs. For each response chunk, Intent Module inference precedes joint audiovisual latent generation by the DiT. The remaining VAE decoding steps for one chunk overlap with Intent Module inference for the following chunk, reducing the interval between successive outputs.

\paragraph{Unified KV Management}
\label{sec:runtime-kv}

A shared KV cache manager tracks cache ownership and device placement, with sliding-window eviction for the Intent Module and periodic history resets for the Realization Module. For each chunk, four denoising forwards read committed historical KV without committing intermediate noisy states. An additional \emph{clean forward} processes the generated latents $z_k^a$ and $z_k^v$ and writes their KV states for subsequent chunks.

At reset boundaries, the manager discards historical media KV before denoising while retaining reference and shared conditioning. The clean forward then establishes a new history, as illustrated by chunk $k+1$ in \figref{fig:runtime}. The reset interval controls retained context, memory use, and context-dependent latency. After a restart, the manager reconstructs the active media cache by replaying completed latents through clean forwards in their original order, following the same reset boundaries and retention policy.

\paragraph{Chunk-Level Overlap and Deadline Scheduling}
\label{sec:runtime-overlap}

Input capture proceeds independently of computation. As shown in \figref{fig:runtime}, the Realization Module uses all 8 GPUs for denoising and the clean forward. Between generation phases, GPUs~0-3 run Intent Module inference for chunk $k+1$, while GPUs~4-7 decode chunk $k$. Generation resumes once both groups finish. After temporal alignment, each 17-frame chunk is released incrementally as one 1-frame anchor followed by four 4-frame packets, each synchronized with its audio.

Let $t_{\mathrm{gen},k}$ include four denoising forwards and the clean forward, $t_{\mathrm{decode},k}^{\mathrm{full}}$ denote complete chunk decoding, $t_{\mathrm{intent},k+1}$ denote the next intent inference, and $t_{\mathrm{switch},k}$ denote additional transition overhead. When the next input context is available and decoding starts together with intent inference, the cycle time is
\begin{equation}
T_{\mathrm{cycle},k}
= t_{\mathrm{gen},k}
+ \max\!\left\{
t_{\mathrm{intent},k+1},
t_{\mathrm{decode},k}^{\mathrm{full}}
\right\}
+ t_{\mathrm{switch},k}.
\label{eq:runtime-stream-period}
\end{equation}
The target is $T_{\mathrm{cycle},k} \leq 17/24\,\mathrm{s}$, with margin for stage-time variation. Complete decoding enters the throughput budget, whereas only first-packet decoding enters the initial response path. The scheduler reserves decoding slots, bounds output lead over playback, and admits work using KV windows, available memory, and estimated stage durations.

\paragraph{Device-Aware Parallel Execution}
\label{sec:runtime-parallel}

The Intent Module uses four-way tensor parallelism (TP), while the Realization Module selects between TP and sequence parallelism (SP). In our SP configuration, each rank holds a complete weight replica and processes part of the token sequence. TP partitions weights and tensor computation. For chunk $k$, the runtime selects the fastest feasible configuration:
\begin{equation}
\begin{aligned}
&p_k^\star
= \arg\min_{p\in\{\mathrm{SP},\mathrm{TP}\}} T_k(p),\\
&\text{s.t.}\quad M_{k,g}(p)\leq B_{k,g},
\quad g=0,\ldots,7.
\end{aligned}
\label{eq:runtime-placement}
\end{equation}
Here, $T_k(p)$ denotes profiled runtime, $M_{k,g}(p)$ peak memory for weights, retained KV, communication buffers, and temporaries, and $B_{k,g}$ memory remaining after other resident stages. We retain weights when memory permits and prefetch nonresident weights during slack, accounting for transfer and synchronization costs. ASR, input transfers, and packet preparation use pipeline gaps as data dependencies allow.
\section{Evaluation and Showcase}
\label{sec:evaluation}
\label{sec:experiments}

\subsection{Human Evaluation}

\paragraph{Evaluation protocol}
We evaluated the model on 30 real-time duplex interaction cases, each pairing a user's audio-visual input with the model's generated audio-visual reply, replayed side by side on a shared timeline. Each reply is judged the way an end user would experience it, at normal playback speed with sound, against six binary criteria: content responsiveness, language adherence, speech intelligibility, visual identity stability, audio-visual synchronization, and audio-visual emotional consistency. Every criterion has written failure conditions, and no partial credit is given.

\paragraph{Results}
Of 180 checks (30 cases × 6 criteria), 168 passed (93.3\%). The clearest strength is generation fidelity: the model passed all 90 checks on language adherence, speech intelligibility, and visual identity stability. Every reply came in the expected language as complete, intelligible speech, and no case showed a visible identity swap, facial or body malformation, or unexplained change in appearance. Cross-modal coherence is nearly as strong. Vocal emotion was consistent with facial expression and body language in 96.7\% of cases, and lip movement stayed aligned with speech in 90.0\%.

\paragraph{Where headroom remains}
Content responsiveness, which asks whether a reply substantively addresses what the user has just said and shown, reached 73.3\% (22/30). It is the most demanding criterion in the protocol, since a duplex model must settle on a reply while speech and video are still streaming in. The misses fall mainly into two patterns. The first is the long tail of conversation, such as jokes and odd remarks, where a fitting reply requires reading intent beyond the literal words, and doing so in real time. The second is intent that only becomes clear when speech and video are read together. In one case, a user sitting indoors said they had just unlocked a shared bike when it started to rain. Given the indoor setting, this reads as a complaint, yet the model treated it as a request for advice and suggested holding off on riding. Both patterns come down to inferring what the user means from everything they say and show, at conversational speed. This is one of the hardest problems in duplex interaction and the central focus of our next iteration, which builds on a foundation that already passed every language, speech, and identity check.

\subsection{Showcase}
\label{sec:showcase}
\label{sec:qualitative-results}

\begin{figure}[H]
  \centering
  \includegraphics[width=\linewidth]{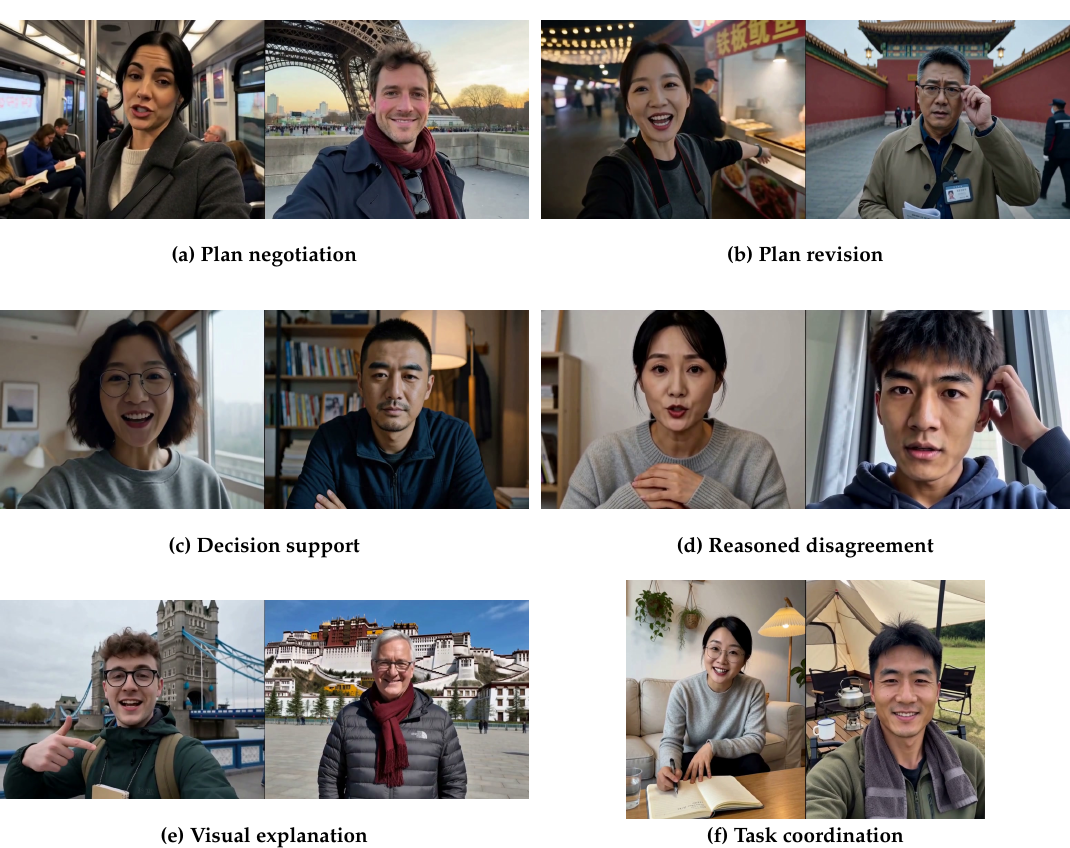}
  \caption{\textbf{Decisions and coordination.} Six additional script-conditioned paired-video examples illustrate how the user and assistant question a proposal, revise a plan, reason through constraints, explain visible evidence, and coordinate a shared task.}
  \label{fig:extended-coordination}
\end{figure}

\begin{figure}[p]
  \centering
  \includegraphics[width=\linewidth]{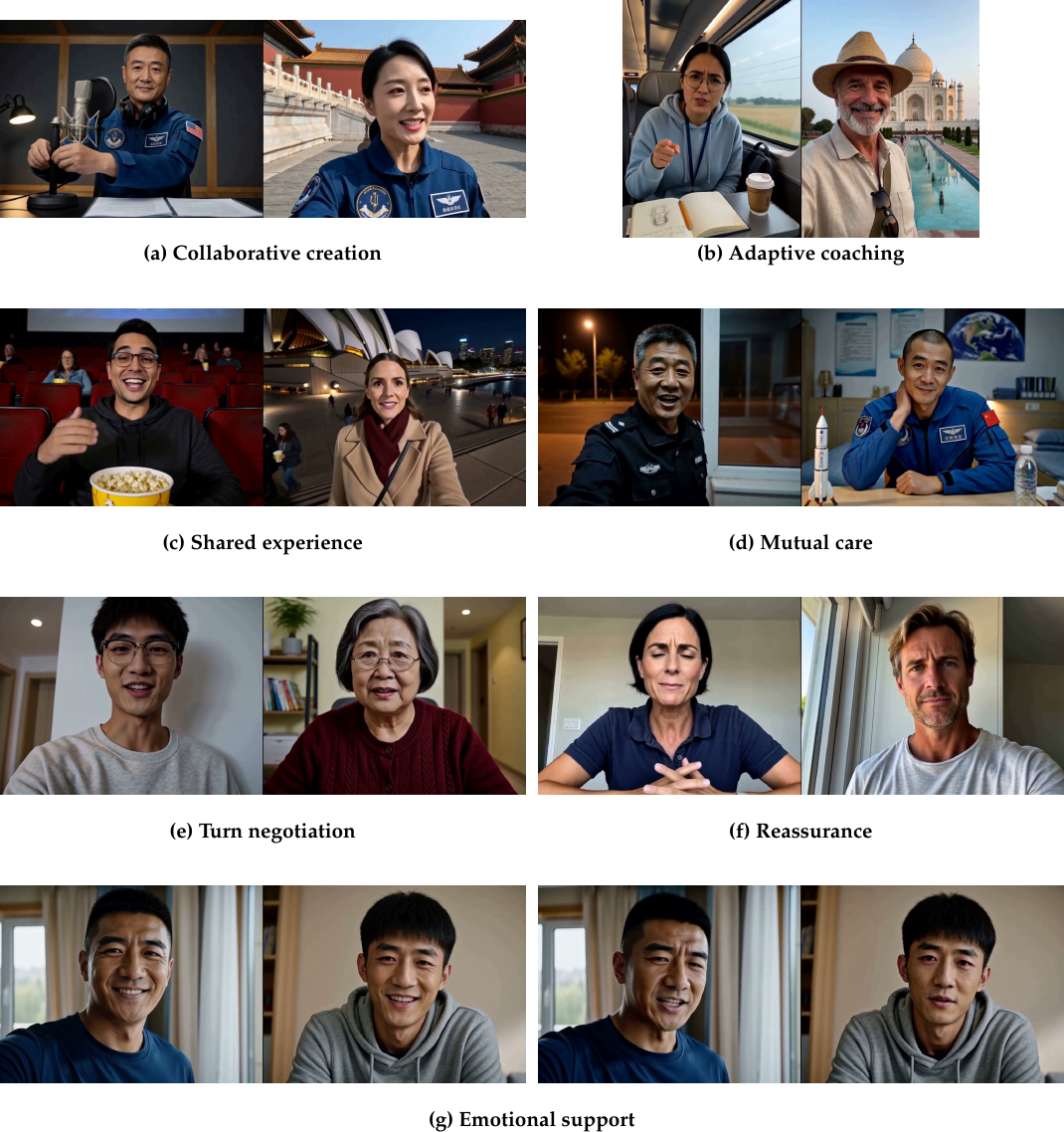}
  \caption{\textbf{Grounded and social responses.} Seven additional script-conditioned conversations connect visible context and dialogue to creative, instructional, and supportive behavior. Panels (a)-(f) show different conversations. Panel (g) shows two moments from one conversation.}
  \label{fig:extended-responses}
\end{figure}

\figref{fig:case-functions} organizes selected examples: plan revision, visual guidance, collaborative design, social support, and interactive customization. These categories connect language to a change in a shared plan, an observable action, a joint solution, or an affective response. The first four categories use script-conditioned training data, the customization example comes from the user demonstration.
\figref{fig:extended-coordination} extends the coverage to negotiation, decision support, explanation, and task coordination. 
\figref{fig:extended-responses} adds collaborative creation, coaching, shared experience, turn negotiation, and supportive responses.

\section{Conclusion}
\label{sec:conclusion}

We introduced the \wrm{} and its initial implementation, \methodname{}, which couples a multimodal Intent Module with a causal audiovisual Realization Module. This report describes the architecture, training procedure, and runtime scheduling used to generate responses while continuing to receive user input. \methodname{} remains preliminary and currently supports basic audiovisual conversation. The examples illustrate this functionality, while response timing, consistency over longer sessions, and recovery from misunderstandings still require broader evaluation. Development is ongoing, and an updated version will follow soon.

\newgeometry{left=1.7cm,right=1.7cm,top=1.5cm,bottom=1.6cm}
\begingroup
\setlength{\parskip}{0pt}
\setlength{\intextsep}{0pt}
\captionsetup{font=small,labelfont=bf,justification=raggedright,singlelinecheck=false,skip=6pt}
\phantomsection\label{sec:front-cases}
\begin{figure}[H]
  \centering
  \includegraphics[width=\linewidth,height=\dimexpr\textheight-45pt\relax,keepaspectratio]{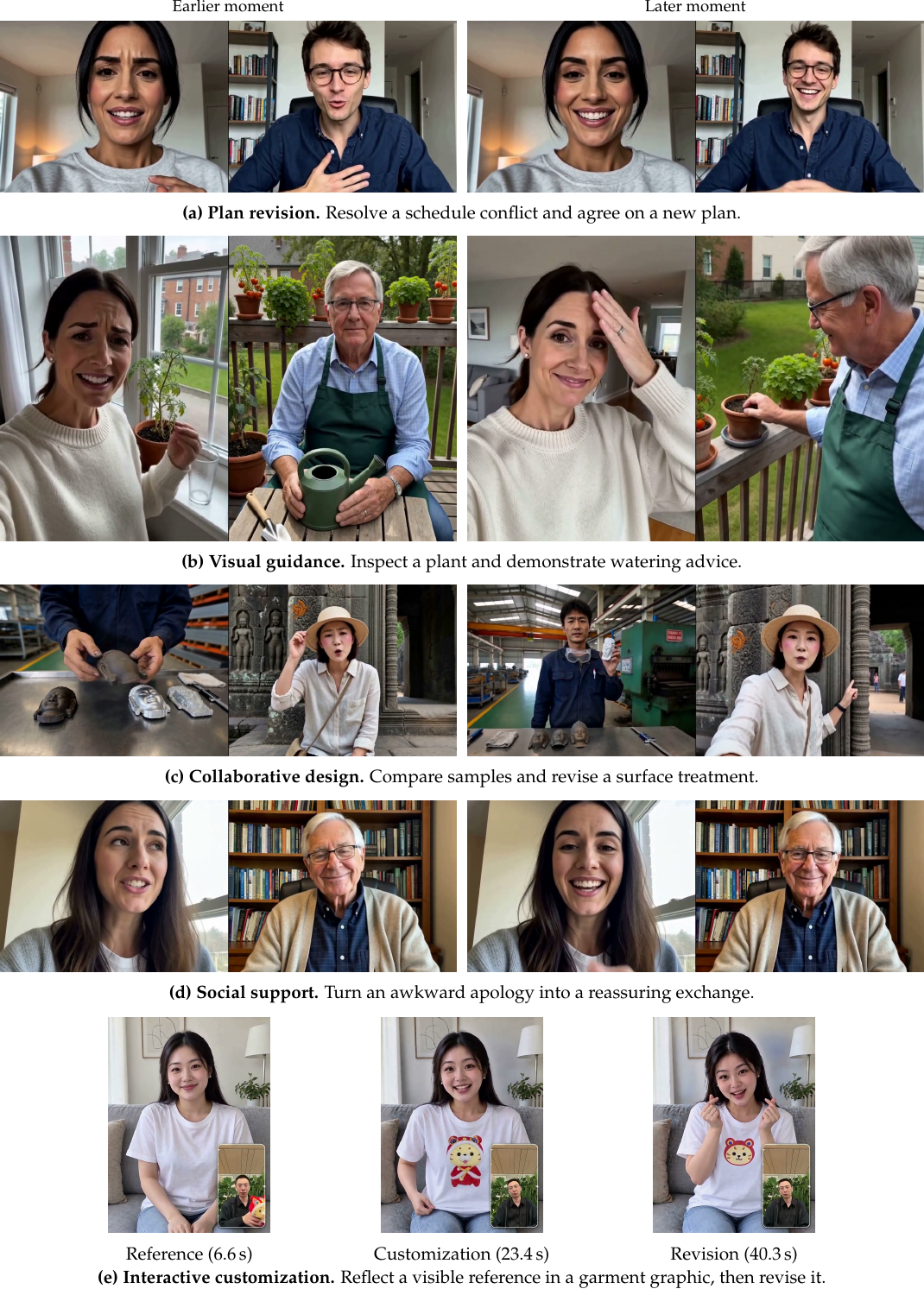}
  \caption{\textbf{Five functions of user--assistant interaction.}
  Examples connect dialogue to plans, visible actions, shared solutions, social responses, and interactive customization. Panels (a)-(d) show offline, script-conditioned paired-video data. Panel (e) shows successive moments from a provided project demo.}
  \label{fig:case-functions}
\end{figure}
\clearpage
\endgroup
\restoregeometry

\FloatBarrier

\bibliographystyle{antgroup}
\bibliography{ref/reference}

@article{ha2018worldmodels,
  author = {Ha, David and Schmidhuber, J{\"u}rgen},
  title = {World Models},
  journal = {arXiv preprint arXiv:1803.10122},
  year = {2018},
  doi = {10.48550/arXiv.1803.10122},
  url = {https://arxiv.org/abs/1803.10122},
}

@article{hafner2025dreamer,
  author = {Hafner, Danijar and Pasukonis, Jurgis and Ba, Jimmy and Lillicrap, Timothy},
  title = {Mastering Diverse Control Tasks through World Models},
  journal = {Nature},
  volume = {640},
  pages = {647--653},
  year = {2025},
  doi = {10.1038/s41586-025-08744-2},
  url = {https://www.nature.com/articles/s41586-025-08744-2},
}

@article{bruce2024genie,
  author = {Bruce, Jake and Dennis, Michael and Edwards, Ashley and Parker-Holder, Jack and Shi, Yuge and Hughes, Edward and Lai, Matthew and Mavalankar, Aditi and Steigerwald, Richie and Apps, Chris and Aytar, Yusuf and Bechtle, Sarah and Behbahani, Feryal and Chan, Stephanie and Heess, Nicolas and Gonzalez, Lucy and Osindero, Simon and Ozair, Sherjil and Reed, Scott and Zhang, Jingwei and Zolna, Konrad and Clune, Jeff and de Freitas, Nando and Singh, Satinder and Rockt{\"a}schel, Tim},
  title = {{Genie}: Generative Interactive Environments},
  journal = {arXiv preprint arXiv:2402.15391},
  year = {2024},
  doi = {10.48550/arXiv.2402.15391},
  url = {https://arxiv.org/abs/2402.15391},
}

@misc{deepmind2024genie2,
  author = {{{Google DeepMind}}},
  title = {{Genie 2}: A Large-Scale Foundation World Model},
  year = {2024},
  howpublished = {Official research release},
  url = {https://deepmind.google/blog/genie-2-a-large-scale-foundation-world-model/},
  note = {Published December 4, 2024; accessed October 30, 2026},
}

@misc{deepmind2025genie3,
  author = {Parker-Holder, Jack and Fruchter, Shlomi},
  title = {{Genie 3}: A New Frontier for World Models},
  year = {2025},
  howpublished = {Google DeepMind official model release},
  url = {https://deepmind.google/blog/genie-3-a-new-frontier-for-world-models/},
  note = {Published August 5, 2025; accessed October 30, 2026},
}

@article{moshiteam2024moshi,
  author = {D{\'e}fossez, Alexandre and Mazar{\'e}, Laurent and Orsini, Manu and Royer, Am{\'e}lie and P{\'e}rez, Patrick and J{\'e}gou, Herv{\'e} and Grave, Edouard and Zeghidour, Neil},
  title = {{Moshi}: A Speech-Text Foundation Model for Real-Time Dialogue},
  journal = {arXiv preprint arXiv:2410.00037},
  year = {2024},
  doi = {10.48550/arXiv.2410.00037},
  url = {https://arxiv.org/abs/2410.00037},
}

@article{qwen2025qwen3vl,
  author = {Bai, Shuai and Cai, Yuxuan and Chen, Ruizhe and Chen, Keqin and Chen, Xionghui and Cheng, Zesen and others},
  title = {{Qwen3-VL} Technical Report},
  journal = {arXiv preprint arXiv:2511.21631},
  year = {2025},
  doi = {10.48550/arXiv.2511.21631},
  url = {https://arxiv.org/abs/2511.21631},
}

@inproceedings{lipman2023flowmatching,
  author = {Lipman, Yaron and Chen, Ricky T. Q. and Ben-Hamu, Heli and Nickel, Maximilian and Le, Matt},
  title = {Flow Matching for Generative Modeling},
  booktitle = {International Conference on Learning Representations},
  year = {2023},
  doi = {10.48550/arXiv.2210.02747},
  url = {https://arxiv.org/abs/2210.02747},
}

@inproceedings{chen2024diffusionforcing,
  author = {Chen, Boyuan and Marti Monso, Diego and Du, Yilun and Simchowitz, Max and Tedrake, Russ and Sitzmann, Vincent},
  title = {Diffusion Forcing: Next-Token Prediction Meets Full-Sequence Diffusion},
  booktitle = {Advances in Neural Information Processing Systems},
  year = {2024},
  doi = {10.48550/arXiv.2407.01392},
  url = {https://arxiv.org/abs/2407.01392},
}

@inproceedings{song2023consistency,
  author = {Song, Yang and Dhariwal, Prafulla and Chen, Mark and Sutskever, Ilya},
  title = {Consistency Models},
  booktitle = {Proceedings of the 40th International Conference on Machine Learning},
  series = {Proceedings of Machine Learning Research},
  volume = {202},
  pages = {32211--32252},
  year = {2023},
  doi = {10.48550/arXiv.2303.01469},
  url = {https://proceedings.mlr.press/v202/song23a.html},
}

@inproceedings{yin2024dmd,
  author = {Yin, Tianwei and Gharbi, Micha{\"e}l and Zhang, Richard and Shechtman, Eli and Durand, Fr{\'e}do and Freeman, William T. and Park, Taesung},
  title = {One-Step Diffusion with Distribution Matching Distillation},
  booktitle = {Proceedings of the IEEE/CVF Conference on Computer Vision and Pattern Recognition},
  year = {2024},
  doi = {10.48550/arXiv.2311.18828},
  url = {https://arxiv.org/abs/2311.18828},
}

@inproceedings{yin2024dmd2,
  author = {Yin, Tianwei and Gharbi, Micha{\"e}l and Park, Taesung and Zhang, Richard and Shechtman, Eli and Durand, Fr{\'e}do and Freeman, William T.},
  title = {Improved Distribution Matching Distillation for Fast Image Synthesis},
  booktitle = {Advances in Neural Information Processing Systems},
  year = {2024},
  doi = {10.48550/arXiv.2405.14867},
  url = {https://arxiv.org/abs/2405.14867},
}

@inproceedings{yin2025selfforcing,
  author = {Huang, Xun and Li, Zhengqi and He, Guande and Zhou, Mingyuan and Shechtman, Eli},
  title = {Self Forcing: Bridging the Train-Test Gap in Autoregressive Video Diffusion},
  booktitle = {Advances in Neural Information Processing Systems},
  year = {2025},
  doi = {10.48550/arXiv.2506.08009},
  url = {https://arxiv.org/abs/2506.08009},
}

@inproceedings{causvid2025,
  author = {Yin, Tianwei and Zhang, Qiang and Zhang, Richard and Freeman, William T. and Durand, Fr{\'e}do and Shechtman, Eli and Huang, Xun},
  title = {From Slow Bidirectional to Fast Autoregressive Video Diffusion Models},
  booktitle = {Proceedings of the IEEE/CVF Conference on Computer Vision and Pattern Recognition},
  year = {2025},
  doi = {10.48550/arXiv.2412.07772},
  url = {https://openaccess.thecvf.com/content/CVPR2025/papers/Yin_From_Slow_Bidirectional_to_Fast_Autoregressive_Video_Diffusion_Models_CVPR_2025_paper.pdf},
}

@inproceedings{wang2025vggt,
  author = {Wang, Jianyuan and Chen, Minghao and Karaev, Nikita and Vedaldi, Andrea and Rupprecht, Christian and Novotny, David},
  title = {{VGGT}: Visual Geometry Grounded Transformer},
  booktitle = {Proceedings of the IEEE/CVF Conference on Computer Vision and Pattern Recognition},
  year = {2025},
  doi = {10.48550/arXiv.2503.11651},
  url = {https://arxiv.org/abs/2503.11651},
}

@misc{h3modelcard,
  author = {{{MiniMax}}},
  title = {{MiniMax H3}: Official Model Card},
  year = {2026},
  howpublished = {Hugging Face model card},
  url = {https://huggingface.co/MiniMaxAI/MiniMax-H3},
  note = {Accessed October 30, 2026; year denotes the access year of the undated model card},
}

@article{robbyant2026lingbot,
  author = {{Robbyant Team}},
  title = {Advancing Open-source World Models},
  journal = {arXiv preprint arXiv:2601.20540},
  year = {2026},
  doi = {10.48550/arXiv.2601.20540},
  url = {https://arxiv.org/abs/2601.20540},
}

@article{huang2026wanstreamer,
  author = {Huang, Lianghua and Wu, Zhi-Fan and Wang, Wei and others},
  title = {{Wan-Streamer v0.1}: End-to-end Real-time Interactive Foundation Models},
  journal = {arXiv preprint arXiv:2606.25041},
  year = {2026},
  doi = {10.48550/arXiv.2606.25041},
  url = {https://arxiv.org/abs/2606.25041},
}

@article{huang2026solarwm,
  author = {Huang, Junchao and Fang, Guian and Qian, Shengju and others},
  title = {{SolarWM}: Open Data and Scalable Training for Long-Horizon Video World Models},
  journal = {arXiv preprint arXiv:2609.02886},
  year = {2026},
  doi = {10.48550/arXiv.2609.02886},
  url = {https://arxiv.org/abs/2609.02886},
}

@article{hu2023gaia,
  author = {Hu, Anthony and Russell, Lloyd and Yeo, Hudson and others},
  title = {{GAIA-1: A Generative World Model for Autonomous Driving}},
  journal = {arXiv preprint arXiv:2309.17080},
  year = {2023},
  doi = {10.48550/arXiv.2309.17080},
  url = {https://arxiv.org/abs/2309.17080},
}

@article{agarwal2025cosmos,
  author = {{NVIDIA}},
  title = {{Cosmos World Foundation Model Platform for Physical AI}},
  journal = {arXiv preprint arXiv:2501.03575},
  year = {2025},
  doi = {10.48550/arXiv.2501.03575},
  url = {https://arxiv.org/abs/2501.03575},
}

@article{wang2026ellsa2,
  author = {Wang, Siyin and Yu, Wenyi and Chiba, Terumi and others},
  title = {{Full-Duplex Interaction with Speech, Expressions and Gestures}},
  journal = {arXiv preprint arXiv:2610.21452},
  year = {2026},
  doi = {10.48550/arXiv.2610.21452},
  url = {https://arxiv.org/abs/2610.21452},
}

@article{xu2025qwen25omni,
  author = {Xu, Jin and Guo, Zhifang and He, Jinzheng and others},
  title = {{Qwen2.5-Omni Technical Report}},
  journal = {arXiv preprint arXiv:2503.20215},
  year = {2025},
  doi = {10.48550/arXiv.2503.20215},
  url = {https://arxiv.org/abs/2503.20215},
}

@article{xu2025qwen3omni,
  author = {Xu, Jin and Guo, Zhifang and Hu, Hangrui and others},
  title = {{Qwen3-Omni Technical Report}},
  journal = {arXiv preprint arXiv:2509.17765},
  year = {2025},
  doi = {10.48550/arXiv.2509.17765},
  url = {https://arxiv.org/abs/2509.17765},
}

@article{tian2024emo,
  author = {Tian, Linrui and Wang, Qi and Zhang, Bang and Bo, Liefeng},
  title = {{EMO: Emote Portrait Alive -- Generating Expressive Portrait Videos with Audio2Video Diffusion Model under Weak Conditions}},
  journal = {arXiv preprint arXiv:2402.17485},
  year = {2024},
  doi = {10.48550/arXiv.2402.17485},
  url = {https://arxiv.org/abs/2402.17485},
}

@article{lin2025omnihuman,
  author = {Lin, Gaojie and Jiang, Jianwen and Yang, Jiaqi and others},
  title = {{OmniHuman-1: Rethinking the Scaling-Up of One-Stage Conditioned Human Animation Models}},
  journal = {arXiv preprint arXiv:2502.01061},
  year = {2025},
  doi = {10.48550/arXiv.2502.01061},
  url = {https://arxiv.org/abs/2502.01061},
}

@article{chern2025livetalk,
  author = {Chern, Ethan and Hu, Zhulin and Tang, Bohao and others},
  title = {{LiveTalk: Real-Time Multimodal Interactive Video Diffusion via Improved On-Policy Distillation}},
  journal = {arXiv preprint arXiv:2512.23576},
  year = {2025},
  doi = {10.48550/arXiv.2512.23576},
  url = {https://arxiv.org/abs/2512.23576},
}

@article{low2025ovi,
  author = {Low, Chetwin and Wang, Weimin and Katyal, Calder},
  title = {{Ovi: Twin Backbone Cross-Modal Fusion for Audio-Video Generation}},
  journal = {arXiv preprint arXiv:2510.01284},
  year = {2025},
  doi = {10.48550/arXiv.2510.01284},
  url = {https://arxiv.org/abs/2510.01284},
}

@article{openmoss2026mova,
  author = {{SII-OpenMOSS Team}},
  title = {{MOVA: Towards Scalable and Synchronized Video-Audio Generation}},
  journal = {arXiv preprint arXiv:2602.08794},
  year = {2026},
  doi = {10.48550/arXiv.2602.08794},
  url = {https://arxiv.org/abs/2602.08794},
}

@article{hacohen2026ltx2,
  author = {HaCohen, Yoav and Brazowski, Benny and Chiprut, Nisan and others},
  title = {{LTX-2: Efficient Joint Audio-Visual Foundation Model}},
  journal = {arXiv preprint arXiv:2601.03233},
  year = {2026},
  doi = {10.48550/arXiv.2601.03233},
  url = {https://arxiv.org/abs/2601.03233},
}

@article{zhu2026omniliveavatar,
  author = {Zhu, Lunjie and Ge, Xingtong and Lin, Fangyu and others},
  title = {{Omni-LiveAvatar: Minute-Level Real-Time Streaming Joint Audio-Video Avatar Generation}},
  journal = {arXiv preprint arXiv:2608.13602},
  year = {2026},
  doi = {10.48550/arXiv.2608.13602},
  url = {https://arxiv.org/abs/2608.13602},
}

@article{qwen2026asr,
  author = {Shi, Xian and Wang, Xiong and Guo, Zhifang and others},
  title = {{Qwen3-ASR Technical Report}},
  journal = {arXiv preprint arXiv:2601.21337},
  year = {2026},
  doi = {10.48550/arXiv.2601.21337},
  url = {https://arxiv.org/abs/2601.21337},
}

@article{su2026omniforcing,
  author = {Su, Yaofeng and Li, Yuming and Xue, Zeyue and others},
  title = {{OmniForcing: Unleashing Real-time Joint Audio-Visual Generation}},
  journal = {arXiv preprint arXiv:2603.11647},
  year = {2026},
  doi = {10.48550/arXiv.2603.11647},
  url = {https://arxiv.org/abs/2603.11647},
}

@article{zhu2026causalforcing,
  author = {Zhu, Hongzhou and Zhao, Min and He, Guande and others},
  title = {{Causal Forcing: Autoregressive Diffusion Distillation Done Right for High-Quality Real-Time Interactive Video Generation}},
  journal = {arXiv preprint arXiv:2602.02214},
  year = {2026},
  doi = {10.48550/arXiv.2602.02214},
  url = {https://arxiv.org/abs/2602.02214},
}

@article{zhao2026causalforcingpp,
  author = {Zhao, Min and Zhu, Hongzhou and Zheng, Kaiwen and others},
  title = {{Causal Forcing++: Scalable Few-Step Autoregressive Diffusion Distillation for Real-Time Interactive Video Generation}},
  journal = {arXiv preprint arXiv:2605.15141},
  year = {2026},
  doi = {10.48550/arXiv.2605.15141},
  url = {https://arxiv.org/abs/2605.15141},
}

@misc{sileroteam2024vad,
  author = {{Silero Team}},
  title = {Silero VAD: Pre-trained Enterprise-grade Voice Activity Detector ({VAD}), Number Detector and Language Classifier},
  year = {2024},
  howpublished = {GitHub software repository},
  url = {https://github.com/snakers4/silero-vad},
  note = {Software citation follows the official CITATION.cff; accessed November 7, 2026},
}

@inproceedings{gao2023funasr,
  author = {Gao, Zhifu and Li, Zerui and Wang, Jiaming and Luo, Haoneng and Shi, Xian and Chen, Mengzhe and Li, Yabin and Zuo, Lingyun and Du, Zhihao and Zhang, Shiliang},
  title = {FunASR: A Fundamental End-to-End Speech Recognition Toolkit},
  booktitle = {Interspeech 2023},
  pages = {1593--1597},
  year = {2023},
  doi = {10.21437/Interspeech.2023-1428},
  url = {https://www.isca-archive.org/interspeech_2023/gao23g_interspeech.html},
}

@inproceedings{gao2022paraformer,
  author = {Gao, Zhifu and Zhang, Shiliang and McLoughlin, Ian and Yan, Zhijie},
  title = {Paraformer: Fast and Accurate Parallel Transformer for Non-autoregressive End-to-End Speech Recognition},
  booktitle = {Interspeech 2022},
  pages = {2063--2067},
  year = {2022},
  doi = {10.21437/Interspeech.2022-9996},
  url = {https://www.isca-archive.org/interspeech_2022/gao22b_interspeech.html},
}

@inproceedings{wang2023campplus,
  author = {Wang, Hui and Zheng, Siqi and Chen, Yafeng and Cheng, Luyao and Chen, Qian},
  title = {CAM++: A Fast and Efficient Network for Speaker Verification Using Context-Aware Masking},
  booktitle = {Interspeech 2023},
  pages = {5301--5305},
  year = {2023},
  doi = {10.21437/Interspeech.2023-1513},
  url = {https://www.isca-archive.org/interspeech_2023/wang23ha_interspeech.html},
}

@article{lingbot-world-v2,
      title={Infinite Worlds with Versatile Interactions}, 
      author={Zelin Gao and Qiuyu Wang and Jiapeng Zhu and Jingye Chen and Zichen Liu and Qingyan Bai and Jiahao Wang and Yufeng Yuan and Hanlin Wang and Yichong Lu and Ka Leong Cheng and Haojie Zhang and Jian Gao and Tianrui Feng and Yuzheng Liu and Yao Yao and Yinghao Xu and Xing Zhu and Yujun Shen and Hao Ouyang},
      journal={arXiv preprint arXiv:2607.07534},
      year={2026}
}

@article{chen2026hy,
  title={Hy-Nomas: Towards Free-form Interactive World Generation},
  author={Chen, Qiaoling and Cheng, Jiaxiang and Deng, Xueqing and Hang, Tiankai and Hou, Ji and Li, Jiaqi and Liu, Yifan and Lu, Tao and Tang, Junshu and Tang, Yu and others},
  journal={arXiv preprint arXiv:2610.36729},
  year={2026}
}

@article{brooks2024sora,
  title={Video generation models as world simulators},
  author={Tim Brooks and Bill Peebles and Connor Holmes and Will DePue and Yufei Guo and Li Jing and David Schnurr and Joe Taylor and Troy Luhman and Eric Luhman and Clarence Ng and Ricky Wang and Aditya Ramesh},
  year={2024},
  url={https://openai.com/research/video-generation-models-as-world-simulators},
}

@misc{valevski2025gamengen,
      title={Diffusion Models Are Real-Time Game Engines}, 
      author={Dani Valevski and Yaniv Leviathan and Moab Arar and Shlomi Fruchter},
      year={2025},
      eprint={2408.14837},
      archivePrefix={arXiv},
      primaryClass={cs.LG},
      url={https://arxiv.org/abs/2408.14837}, 
}
\end{document}